\documentclass[11pt]{article}

\usepackage{acl}
\usepackage{helvet}
\usepackage{times}
\usepackage{latexsym}
\usepackage[T1]{fontenc}
\usepackage[utf8]{inputenc}
\usepackage{microtype}
\usepackage{newunicodechar}
\newunicodechar{├}{\textbar\textemdash}
\usepackage{booktabs}   
\usepackage{tcolorbox}
\usepackage{colortbl}   
\usepackage{array}      
\usepackage{pifont}     
\usepackage{amssymb}    
\usepackage{graphicx}   
\usepackage{algorithm}
\usepackage{algorithmic}
\usepackage[utf8]{inputenc}
\usepackage{enumitem} 
\usepackage{amsmath}  
\usepackage{xcolor}   
\usepackage{graphicx} 
\usepackage{multirow}
\usepackage{subcaption}  
\usepackage{arydshln}
\usepackage{listings}
\usepackage[frozencache=true,cachedir=minted-cache]{minted}
\usepackage{algorithm}
\usepackage{framed}
\newcommand{\Func}[1]{\textsc{#1}}
\newcommand{\LLM}[1]{\textsc{LLM}_{\text{#1}}}
\usepackage{makecell}

\AtBeginDocument{
    \numberwithin{listing}{section}
    \numberwithin{lstlisting}{section}
}
\definecolor{cGreen}{RGB}{0,150,0}
\definecolor{cRed}{RGB}{200,0,0}
\definecolor{cBlue}{RGB}{0,0,200}
\definecolor{bgGray}{RGB}{245,245,245}
\definecolor{shadecolor}{gray}{0.95}
\tcbuselibrary{skins, breakable}
\definecolor{bg_gray}{RGB}{245,245,245}
\definecolor{c_red}{RGB}{220,53,69}
\definecolor{c_green}{RGB}{40,167,69}
\definecolor{c_blue}{RGB}{0,123,255}
\definecolor{c_purple}{RGB}{111,66,193}

\definecolor{lightblue}{RGB}{222,235,247}
\definecolor{correctgreen}{RGB}{0,153,0}
\definecolor{wrongred}{RGB}{204,0,0}
\definecolor{codebg}{RGB}{245,245,245}
\definecolor{forestgreen}{RGB}{0, 153, 76}
\definecolor{brickred}{RGB}{204, 0, 0}
\definecolor{amber}{RGB}{255, 191, 0}
\definecolor{lightteal}{RGB}{225, 240, 240} 
\definecolor{bg-red}{HTML}{F4CCCC}    
\definecolor{bg-orange}{HTML}{FCE5CD} 
\definecolor{bg-yellow}{HTML}{FFF2CC} 
\definecolor{bg-lgreen}{HTML}{D9EAD3} 
\definecolor{bg-mgreen}{HTML}{B6D7A8} 
\definecolor{bg-dgreen}{HTML}{93C47D} 
\definecolor{good}{RGB}{0,120,60}
\definecolor{bad}{RGB}{180,30,30}
\newcommand{\cmark}{\textcolor{forestgreen}{\scalebox{1.2}{\ding{51}}}}
\newcommand{\strong}{\cmark}
\newcommand{\xmark}{\textcolor{brickred}{\scalebox{1.2}{\ding{55}}}}
\newcommand{\tmark}{\textcolor{amber}{\scalebox{1.2}{$\blacktriangle$}}}
\newcolumntype{a}{>{\columncolor{lightteal}}c}

\title{ASTRA: Adaptive Semantic Tree Reasoning Architecture for Complex Table Question Answering}

\author{
    Xiaoke Guo$^1$, Songze Li$^1$, Zhiqiang Liu$^1$, Zhaoyan Gong$^1$, \\
    \textbf{Yuanxiang Liu$^1$, Huajun Chen$^1$, Wen Zhang$^{1,}$\thanks{ Corresponding author.}} \\
    $^1$Zhejiang University \\
    \texttt{\{guoxiaoke, zhang.wen\}@zju.edu.cn}
}
\begin{document}
\maketitle

\begin{abstract}

Table serialization remains a critical bottleneck for Large Language Models (LLMs) in complex table question answering, hindered by challenges such as structural neglect, representation gaps, and reasoning opacity. Existing serialization methods fail to capture explicit hierarchies and lack schema flexibility, while current tree-based approaches suffer from limited semantic adaptability. To address these limitations, we propose \textbf{ASTRA} (\textbf{A}daptive \textbf{S}emantic \textbf{T}ree \textbf{R}easoning \textbf{A}rchitecture) including two main modules, \textbf{AdaSTR} and \textbf{DuTR}. First, we introduce \textbf{AdaSTR}, which leverages the global semantic awareness of LLMs to reconstruct tables into Logical Semantic Trees. This serialization explicitly models hierarchical dependencies and employs an adaptive mechanism to optimize construction strategies based on table scale. Second, building on this structure, we present \textbf{DuTR}, a dual-mode reasoning framework that integrates tree-search-based textual navigation for linguistic alignment and symbolic code execution for precise verification. Experiments on complex table benchmarks demonstrate that our method achieves state-of-the-art (SOTA) performance. Our code is available at \url{https://github.com/zjukg/ASTRA}.

\end{abstract}

\section{Introduction}

Large Language Models (LLMs) have achieved remarkable success across a wide spectrum of natural language tasks. As a fundamental format for data storage and presentation, tables—particularly complex tables characterized by hierarchical headers and merged cells—are ubiquitous in high-value domains. Consequently, the community has devoted significant efforts to enhance the tabular reasoning and question-answering capabilities of LLMs.

Recent studies~\cite{fang2024largelanguagemodelsllmstabular,sui-etal-2024-tap4llm} have identified table serialization—the process of converting structured tables into sequential representations compatible with LLM inputs—as a critical bottleneck for Table Question Answering (TableQA). Specifically, we observe that current serialization methods face four major challenges when handling complex tables: (1) \textbf{Structural Neglect:} Complex tables often contain intricate layouts with hierarchical relationships and semantic dependencies embedded in their structure, which LLMs frequently overlook. (2) \textbf{Representation Gap:} The intrinsic representational mismatch between two-dimensional structured tables and the one-dimensional sequential nature of LLMs hinders the precise localization of fine-grained evidence, which is important for reasoning over complex tables. (3) \textbf{Reasoning Opacity:} Existing methods directly employ LLMs for numerical computation over tables, which function as a black box lacking interpretable execution traces, leading to unverified numerical hallucinations. (4) \textbf{Schema Inflexibility:} Complex tables exhibit significant structural diversity. Existing table parsing strategies that rely on rigid, rule-based approaches fail to adapt to irregular layouts, limiting generalization.

Despite the variety of existing approaches, none address these challenges simultaneously: First, graph-based approaches like GraphOTTER~\cite{li-etal-2025-graphotter} atomize tables into triples. While this strategy mitigates schema inflexibility, it still fails to explicitly represent the hierarchical structures and semantic information in complex tables. Conversely, methods such as RelationalCoder~\cite{dong-etal-2025-relationalcoder} attempt to convert complex tables into relational formats. Although these approaches can elucidate semantic associations among headers via schemas and facilitate the integration of Text-to-SQL methodologies~\cite{wang2024chainoftableevolvingtablesreasoning}, they frequently introduce data redundancy or sparsity when applied to asymmetric structures. For instance, a `Laptop' sub-table may contain a `GPU' attribute, whereas a `Mouse' sub-table lacks `GPU' but includes `DPI'; forced integration into a unified schema induces data sparsity, while split storage introduces data redundancy. Furthermore, while current tree-based solutions like ST-Raptor~\cite{tang2025straptorllmpoweredsemistructuredtable} demonstrate the utility of hierarchical representations, their reliance on rule-based construction renders them fragile (details are discussed in Appendix \ref{appendix:difference}). Crucially, they fail to discern between physical hierarchy and semantic associations. 

To achieve a more effective serialization that enhances LLMs' reasoning capabilities over complex tables, we propose \textbf{Ada}ptive \textbf{S}emantic \textbf{T}ree \textbf{R}econstruction (\textbf{AdaSTR}), as illustrated in Table~\ref{tab:transposed}, which simultaneously addresses the challenges by satisfying the following requirements:

\begin{figure}[t]
    \centering
    \includegraphics[width=1\linewidth]{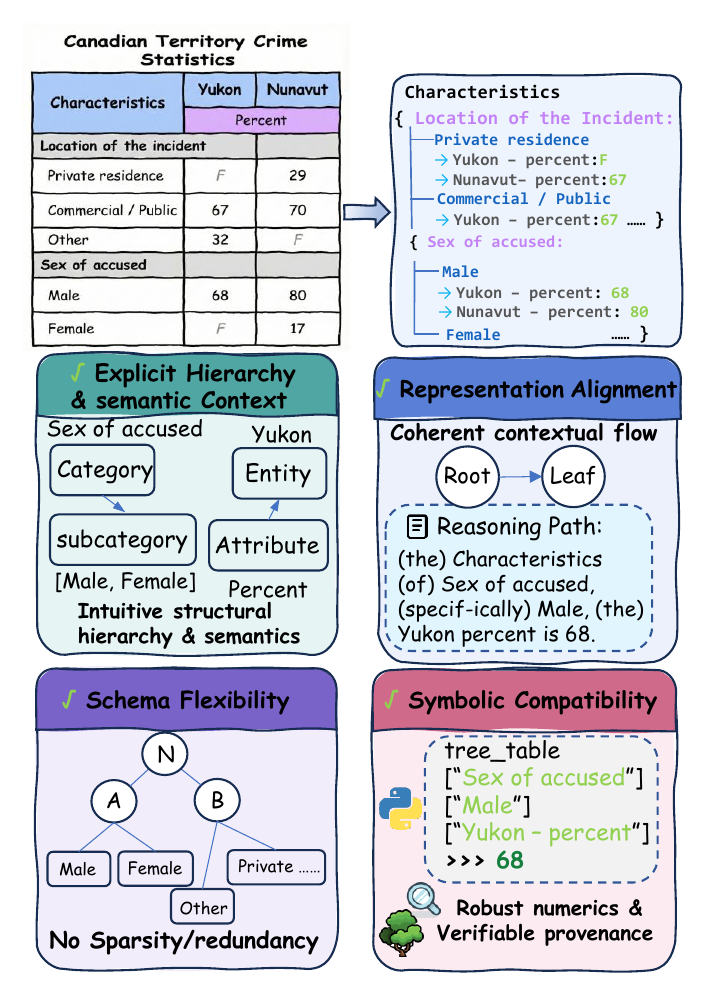} 
    \vspace{-9mm}  
    \caption{Key desiderata for robust table serialization and their instantiation in AdaSTR.}
    \label{fig:requirments}
    \vspace{-4mm}
\end{figure}

\paragraph{Explicit Hierarchy \& Semantic Context:} 
To resolve \textit{Structural Neglect}, we make the serialization preserve the explicit hierarchical structure inherent in table headers (e.g., taxonomic levels like $\text{Category} \rightarrow \text{Subcategory}$), while simultaneously revealing the relational implicit semantic dependencies within the data dimension (e.g., $\text{Entity} \leftrightarrow \text{Attribute}$ mappings). This dual focus enables a more precise representation of the complex table's global semantics.

\paragraph{Representation Alignment:} 
To mitigate \textit{Representation Gap}, we design the serialization to bridge the 2D-to-1D modality gap by expressing tabular content as coherent, natural-language-like sequences that align with the distribution of LLM's pre-training data~\cite{wu2025tabulardataunderstandingllms}. Such representation alignment not only enhances comprehension for LLMs but also facilitates the retrieval of critical content relevant to specific queries.

\paragraph{Symbolic Compatibility:} 
To tackle \textit{Reasoning Opacity}, we structure the serialization to support \emph{code-based} symbolic reasoning. This not only mitigates LLMs' weaknesses in numerical computations~\cite{liu2023rethinkingtabulardataunderstanding,zhang-etal-2024-syntqa} but also ensures deterministic and verifiable data provenance, allowing users to validate exact data sources.

\paragraph{Schema Flexibility:} 
To overcome \textit{Schema Inflexibility}, we optimize the serialization to avoid introducing irrelevant redundancy that exacerbates the model's inference burden, ensuring robust adaptability across arbitrary table formats.

\begin{table}[t]
\centering
\small 
\resizebox{\columnwidth}{!}{
    \renewcommand{\arraystretch}{1.4} 
    \begin{tabular}{l c c c c c} 
    \toprule
    \textbf{Method} & 
    \textbf{\begin{tabular}[c]{@{}c@{}}Explicit\\ Hierarchy\end{tabular}} & 
    \textbf{\begin{tabular}[c]{@{}c@{}}Semantic\\ Context\end{tabular}} & 
    \textbf{\begin{tabular}[c]{@{}c@{}}Representation\\ Alignment\end{tabular}} & 
    \textbf{\begin{tabular}[c]{@{}c@{}}Schema\\ Flexibility\end{tabular}} & 
    \textbf{\begin{tabular}[c]{@{}c@{}}Symbolic\\ Compatibility\end{tabular}} \\ 
    \midrule
    
    Textual Serialization & 
    \tmark & \xmark & \xmark & \tmark & \xmark \\
    \textit{(Html/MarkDown)} & & & & & \\
    
    Triples & 
    \xmark & \xmark & \xmark & \cmark & \tmark \\
    \textit{(GraphOTTER)} & & & & & \\

    Relational & 
    \tmark & \cmark & \tmark & \xmark & \cmark \\
    \textit{(TabFormer)} & & & & & \\

    Physical Tree & 
    \cmark & \xmark & \tmark & \cmark & \cmark \\
    \textit{(ST-Raptor)} & & & & & \\
    
    \midrule
    
    \textbf{Semantic Tree} & 
    \multirow{2}{*}{\strong} & 
    \multirow{2}{*}{\strong} & 
    \multirow{2}{*}{\strong} & 
    \multirow{2}{*}{\strong} & 
    \multirow{2}{*}{\strong} \\
    \textbf{(Ours)} & & & & & \\
    
    \bottomrule
    \end{tabular}
}
\caption{Comparison of serialization methods (\cmark~Support; \tmark~Partial; \xmark~No Support).}
\label{tab:transposed}
\vspace{-4mm}
\end{table}


As illustrated in Figure~\ref{fig:requirments}, our serialized semantic tree captures \textbf{explicit hierarchy} and \textbf{semantic context} through node-to-node relationships. To effectively leverage this structure, we introduce \textbf{DuTR} (\textbf{Du}al-Mode \textbf{T}ree \textbf{R}easoning). By incorporating tree-search-based textual reasoning, DuTR efficiently navigates the tree structure to locate critical information, where the resulting paths \textbf{with aligned representations} serve as coherent text streams. Furthermore, the tree structure inherently offers \textbf{schema flexibility} and can be stored in Python dictionary format to facilitate interpretable \textbf{symbolic reasoning}. To balance efficiency and accuracy, we also introduce an adaptive mechanism that dynamically selects tree construction strategies based on \textbf{table scale} and \textbf{token density}.

We name our unified method \textbf{ASTRA} (\textbf{A}daptive \textbf{S}emantic \textbf{T}ree \textbf{R}easoning \textbf{A}rchitecture), which integrates the adaptive construction module (\textbf{AdaSTR}) with the reasoning engine (\textbf{DuTR}). Specifically, DuTR treats the tree simultaneously as a navigable text corpus for semantic retrieval and a structured object for symbolic execution, effectively combining the semantic flexibility of linguistic retrieval with the computational precision of symbolic execution. Overall, our main contributions are as follows:

\begin{itemize}[leftmargin=*, label=$\bullet$]

    \item We identify four critical challenges that constitute the central bottleneck in complex TableQA, and distill the corresponding desiderata for an effective serialization strategy to address them.
    
    \item We introduce AdaSTR, which exploits LLMs’ global semantic awareness to transform a table into a semantic-tree representation. Building on this representation, we further propose the DuTR framework, which achieves high accuracy while providing strong interpretability for table QA.
    
    \item Driven by adaptive tree construction and hybrid reasoning, ASTRA achieves \textbf{SOTA performance} on multiple benchmarks, demonstrating superior generalization and robustness across varied and irregular table formats.

\end{itemize}

\section{Related Work}
\subsection{Adapting Tabular Data for LLMs  }

\paragraph{Textual Table Serialization.} 
Methods utilizing Markdown, HTML, or separators~\cite{sui2024tablemeetsllmlarge} serialize tables into token sequences. However, the syntax of these formats deviates from natural linguistic flow, and it is extremely difficult to retrieve relevant information within such structures, hindering the model’s ability to understand large tables.

\paragraph{Structure-Aware Model Adaptation.} 
Beyond naive serialization, model-centric approaches explicitly encode structure via specialized embeddings~\cite{korkmaz-del-rio-chanona-2024-integrating,he-etal-2025-tablelora}, dedicated table encoders~\cite{li2025tablemodalitylargelanguage}, or continued pre-training~\cite{li2023tablegpttabletunedgptdiverse,su2024tablegpt2largemultimodalmodel,zhang2024tablellamaopenlargegeneralist}. However, these methods typically incur significant adaptation overhead when migrating to new backbones.

\paragraph{Intermediate Structural Representations.} 
To enhance table understanding without training, researchers have investigated transforming tables into model-friendly formats to bridge the semantic gap between structured data and unstructured queries~\cite{tang2025llmagentasdataanalystsurvey, Li_2025}.
Approaches range from relational format~\cite{dong-etal-2025-relationalcoder}, which often suffers from data sparsity on asymmetric layouts, to triple-based atomization~\cite{li-etal-2025-graphotter}, which tends to obscure explicit semantic hierarchies. 
More recently, tree-based methods like ST-Raptor~\cite{tang2025straptorllmpoweredsemistructuredtable} have been proposed; however, their reliance on brittle physical layout heuristics limits their robustness against diverse formatting variations.


\subsection{Table Reasoning Paradigms} 
Table reasoning typically follows two paradigms~\cite{liu2023rethinkingtabulardataunderstanding}
: Textual reasoning (End-to-End) directly generates natural-language rationales and answers from the serialized table context, but may incur numerical errors~\cite{wei2023chainofthoughtpromptingelicitsreasoning}; Symbolic  Reasoning (program-aided) generates executable code (e.g., SQL, Python) for precision~\cite{chen2023programthoughtspromptingdisentangling}. However, it can be sensitive to complex table structures (Details are provided in Appendix~\ref{appendix:reasoning_method}). Our Semantic Tree framework unifies these paradigms by serving as a navigable context for semantic retrieval and a structured object for code generation, effectively combining robustness with precision.

\section{Method}
The goal of TableQA is to synthesize an answer $A$ for a natural language query $Q$, grounded in a source table $T$. While prior benchmarks and methods have largely focused on \emph{flat} tables—typically featuring a single header row where each data cell maps uniquely to one header—we focus on the more challenging task of \textbf{Complex Table QA}.

\subsection{Method Overview}
To address the structural challenges inherent in complex table QA, we formulate the problem as a multi-stage reasoning process consisting of structural transformation and reasoning execution.
\paragraph{General Formulation.} 
First, the raw table $T$ is processed by a representation function $\tau$ to yield a model-readable structured view $\tilde{T} = \tau(T)$. We define $\tau$ as a generalized transformation function, which abstracts the complex layout into a machine-friendly format. Subsequently, a model $\mathcal{M}_\theta$ serves as the core reasoning engine within a workflow $\Phi$. The model interacts with the transformed view $\tilde{T}$ and the query $Q$ to derive the final answer:
\begin{equation}
    A = \Phi_{\mathcal{M}}(\tilde{T}, Q)
\end{equation}

\begin{figure*}[t] 
    \centering 
    \includegraphics[width=\linewidth]{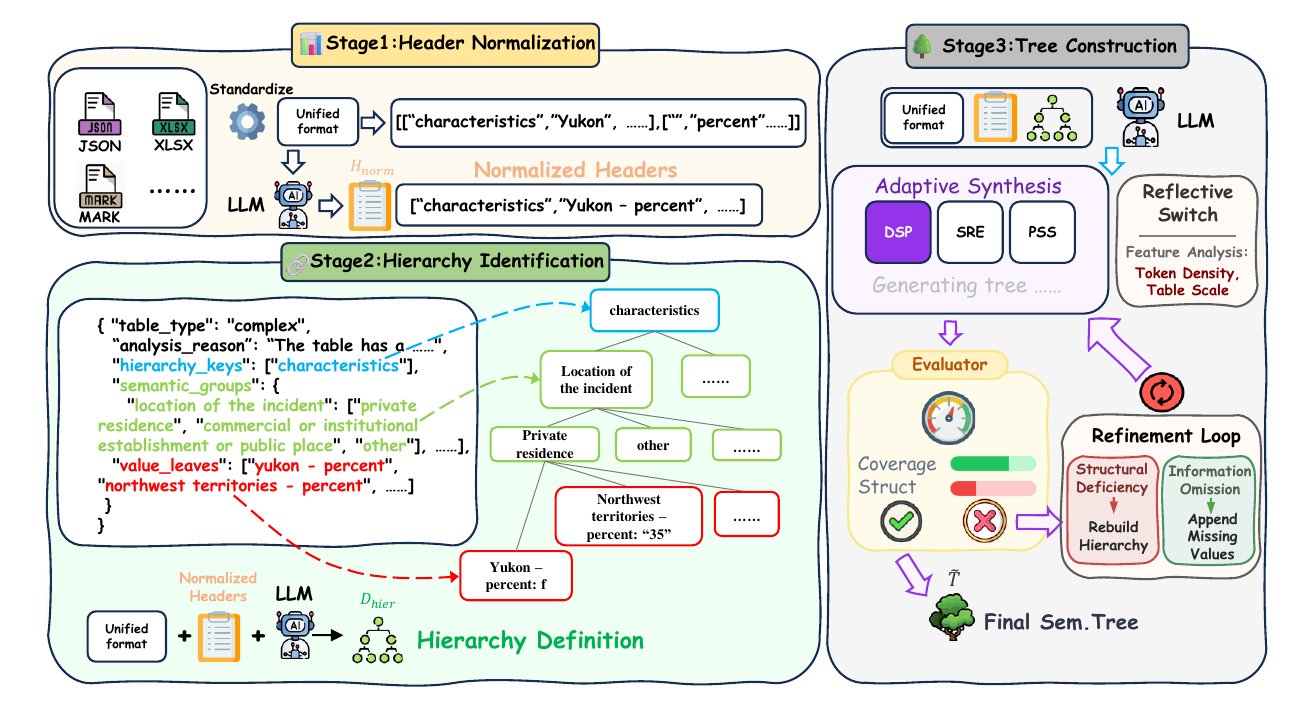} 
    \vspace{-9mm}  
    \caption{Overview of the Adaptive Semantic Tree Reconstruction process.
} 
    \label{fig:tree_construction} 
    \vspace{-3mm}
\end{figure*}

\paragraph{Method Instantiation.} 
Specifically, our method ASTRA comprises two core phases, mapping specifically to $\tau$ and $\Phi$: (1) \textbf{AdaSTR ($\tau \rightarrow$ Table Serialization)}, which transforms a complex table $T$ into a semantic tree $\tilde{T}$; and (2) \textbf{DuTR ($\Phi \rightarrow$ Table Reasoning)}, which integrates innovative tree-based textual reasoning and symbolic code execution, culminating in an answer selection module to determine the optimal final answer.


\subsection{ AdaSTR: Adaptive Semantic Tree Reconstruction}

As illustrated in Figure~\ref{fig:tree_construction}, the AdaSTR framework is orchestrated through three interconnected modules designed to ensure structural robustness and semantic fidelity across heterogeneous tables.

\subsubsection{Semantic Parsing and Schema Detection}
Given a complex table $T$, we first standardize it into a unified format---a flat ordered list structure $l$, 
establishing a reference for the subsequent quality assessment. We then employ a \textbf{Header Identification \& Normalization (HIN)} module. As shown in  Figure~\ref{fig:tree_construction} (stage 1), raw tables often contain sub-headers like \texttt{Percent} that are meaningless without their parent context; \textbf{HIN} resolves this by consolidating vertical dependencies into qualified keys (e.g., merging \texttt{Yukon} with \texttt{Percent} to form \texttt{Yukon-Percent}). This yields a normalized header list $H_{norm}$, which excavates the implicit semantic context, anchoring isolated attributes to their corresponding entities. The subsequent critical phase is \textbf{Hierarchy Identification (HID)}. As depicted in Stage 2 in Figure~\ref{fig:tree_construction}, the \textbf{HID} module harnesses the LLM to mine Hidden Semantic Groups from $H_{norm}$. Continuing the previous example, LLM observes that \texttt{Yukon-Percent} and \texttt{Northwest-Percent} share an identical structure (Region-Metric). Consequently, it abstracts them into a latent high-level concept (e.g., \textit{Regional Statistics}), thereby organizing the extracted semantic units into an explicit hierarchy. Finally, the HID module synthesizes the identified hierarchical structure into a formal representation $D_{hier}$. Crucially, for simple, flat tables lacking deep nesting, the \textbf{Hierarchy Identification (HID)} module and the subsequent adaptive synthesis inherently degenerate to generate a shallow tree. This dynamic fallback mechanism ensures that computational overhead remains strictly proportional to the table's structural complexity, avoiding unnecessary latency for straightforward layouts.

\subsubsection{Adaptive Tree Synthesis Strategies}
This step transforms the table $T$ into a semantic tree $\tilde{T}$ given $D_{hier}$. Considering the heterogeneity of real-world tables---ranging from massive, sparse datasets to text-dense complex tables---a monolithic construction strategy cannot effectively balance efficiency and accuracy. To address this, we devise three adaptive Tree construction modes, with the selection strategy in Appendix~\ref{appendix:mode_selection}.

\paragraph{Direct Semantic Parsing (DSP Mode).}
This mode is optimized for standard complex tables with moderate structural complexity and manageable size. In this setting, the LLM functions as an end-to-end generator, directly outputting the complete semantic tree $\tilde{T}$ based on the parsed schema $D_{hier}$.

\begin{figure*}[t]
    \centering
    \includegraphics[width=\linewidth]{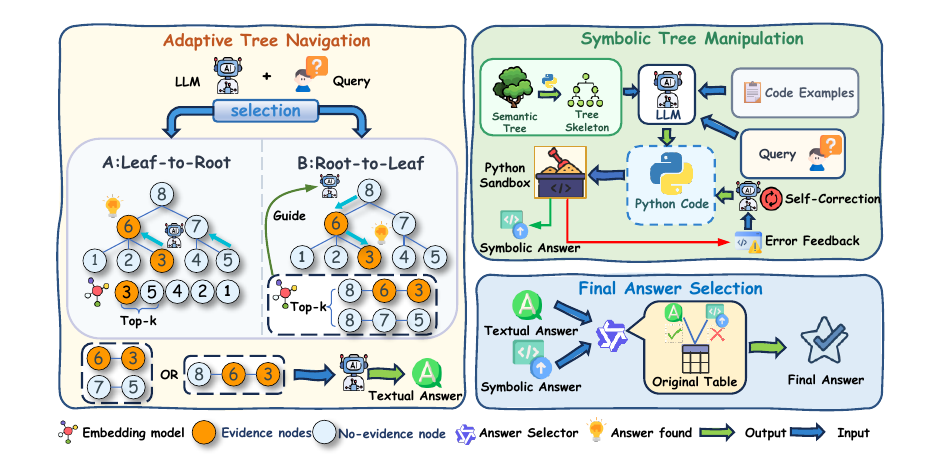} 
    \vspace{-9mm}  
    \caption{Overview of the Dual-Mode Tree Reasoning process.}
    \label{fig:tree_qa}
    \vspace{-3mm}
\end{figure*}

\paragraph{Symbolic Reference Encoding (SRE Mode).} Tailored for verbose-content tables (e.g., financial reports) where generating full textual content is inefficient and risks diluting the model's structural focus, we introduce a Symbolic Placeholder Strategy. Inspired by address-based compressed encodings like SpreadsheetLLM~\cite{dong-etal-2024-encoding}, we employ coordinate abstraction. Guided by the schema $D_{hier}$ and a coordinate-tagged view $T_{grid}$, the LLM is instructed to generate \textit{cell addresses} (e.g., A1 notation like \texttt{C7}) as placeholders within the tree nodes. This yields a compact tree skeleton. Subsequently, a script retrieves the original text via these coordinates to populate the tree $\tilde{T}$. This approach significantly reduces token overhead while preserving structural precision.

\paragraph{Programmatic Structure Synthesis (PSS Mode).} 
Targeting large-scale tables with repetitive substructures, the LLM leverages $D_{hier}$ and $T_{coord}$ to synthesize a loop-based construction script. This script iterates over the coordinate space to instantiate the full tree $\tilde{T}$, ensuring scalability and efficiency.

\subsubsection{Evaluator-Guided Refinement Loop}
To mitigate errors, we employ an \textbf{Evaluator-Guided Refinement Loop}. The Evaluator validates $\tilde{T}$ via: (1) \textbf{Structural Integrity}, which checks path consistency against grid coordinates to prevent logical hallucinations; and (2) \textbf{Information Coverage}, which measures the percentage of mapped cells to verify no semantic information is omitted. If the composite score falls below a threshold, the LLM is instructed to iteratively revise $\tilde{T}$ based on feedback for up to maximum iterations. Implementation details are in Appendix~\ref{appendix:mode_selection}.


\subsection{DuTR: Dual-Mode Tree Reasoning}
As illustrated in Figure~\ref{fig:tree_qa}, we propose a dual-mode reasoning framework grounded in the semantic tree $\tilde{T}$. This structure serves as a foundation for both symbolic manipulation and context-aware retrieval. 

\subsubsection{Adaptive Tree Navigation}

This module analyzes the query to select an optimal traversal strategy (\textit{Leaf-to-Root} or \textit{Root-to-Leaf}), filtering critical paths for the answer. A key advantage of the semantic tree is that it binds each value to its header-path semantics, transforming fragmented cells into context-rich nodes, enabling reliable semantic retrieval on complex tables.

\paragraph{Strategy A: Leaf-to-Root (Algorithm~\ref{alg:leaf2root}).} As illustrated in Panel A of Figure~\ref{fig:tree_qa}, we first filter the top-$k_1$ leaf subset $L_{rel}$ based on query--leaf semantic similarity. Then, it iteratively expands the context depth $d$: at each step, paths associated with $L_{rel}$ are merged into the evidence set $C$. The LLM evaluates whether $C$ is sufficient to derive the answer $A$; if not, the system increments $d$ to include broader parent context until reaching $D_{max}$. This strategy excels in \textbf{aggregation queries}, such as conditional counting of dispersed data points.

\paragraph{Strategy B: Root-to-Leaf (Algorithm~\ref{alg:root2leaf}).} As depicted in Panel B of Figure~\ref{fig:tree_qa}, the system retrieves the top-$k_2$ root-to-leaf paths $P_{topk}$ via semantic similarity, which serve as  global semantic guidance $G$. We then perform a top-down traversal using a stack $S$. At each step, the LLM selects a subset of promising child nodes $N_{sel}$ to expand, conditioned on query $Q$ and guidance $G$. Upon reaching leaf nodes, the path information is accumulated into the evidence set $C$. The LLM dynamically evaluates whether $C$ is sufficient to derive the answer $A$; if affirmative, the traversal triggers early stopping, effectively bypassing the remaining paths. This strategy is well-suited for lookup-style queries requiring precise, localized evidence.

\subsubsection{Symbolic Tree Manipulation}
Pure textual reasoning is prone to error in multi-step queries involving aggregation or complex logical predicates~\cite{ zhang-etal-2024-syntqa}. Thus, as illustrated in Figure~\ref{fig:tree_qa} (top-right), we incorporate symbolic reasoning to ensure computational precision. To accommodate the LLM's context window and focus on logical generation, we abstract the semantic tree into a \textbf{structural skeleton} devoid of verbose values. Furthermore, we construct targeted \textbf{code examples} (e.g., selection, aggregation, and comparison) for tree operations and provide a secure execution environment. If a runtime error occurs, the system enters a \textbf{Self-Correction Loop}, feeding error traces back to the LLM to trigger code regeneration, a failure-aware refinement strategy proven effective in complex structural agents~\cite{liu2026cogcontrollablegraphreasoning, gong2026tempr1unifiedautonomousagent}. This approach unifies accuracy and stability in complex logical reasoning.

\subsection{Final Answer Selection}
As shown in Figure~\ref{fig:tree_qa} (bottom-right), our method yields two candidates: a \textit{Textual Answer} from tree navigation and a \textit{Symbolic Answer} from code execution. To resolve conflicts, we employ a lightweight, open-source LLM as an Answer Selector. When answers differ, it evaluates both against the original table to determine the plausible one. We term this \textbf{Adaptive Selection}. %

\section{Experiment}
\subsection{Experimental Setup}

\paragraph{Datasets.} We evaluate on three complex table benchmarks:
(1) \textbf{AIT-QA}~\cite{katsis-etal-2022-ait}: A domain-specific dataset (airline industry) characterized by deeply nested headers.
(2) \textbf{HiTab}~\cite{cheng-etal-2022-hitab}: A widely used benchmark containing hierarchical tables extracted from statistical reports, emphasizing numerical aggregation.
(3) \textbf{SSTQA}~\cite{tang2025straptorllmpoweredsemistructuredtable}: A dataset introduced by the ST-Raptor framework, included to assess generalization on semi-structured tables.

\paragraph{Baselines.} We compare our method against four categories of baselines:
(1) \textbf{Foundation Models}: \textit{GPT-4o}~\cite{openai2024gpt4technicalreport}, \textit{DeepSeek-V3}~\cite{deepseekai2025deepseekv3technicalreport}, and a recent reasoning-focused model \textit{OpenAI o3}~\cite{openai_o3_o4mini_system_card_2025}. All are prompted via standard textual serialization to benchmark their baseline reasoning performance. 
(2) \textbf{Prompting/Tool-augmented Methods}: including \textit{EEDP}~\cite{srivastava2025evaluatingllmsmathematicalreasoning}, which relies on structured prompting, and \textit{E5}~\cite{zhang-etal-2024-e5}, which adopts an agentic workflow;
(3) \textbf{Table-Specific Adapted Models}: \textit{TableGPT2}~\cite{su2024tablegpt2largemultimodalmodel} and \textit{TableLlama}~\cite{zhang2024tablellamaopenlargegeneralist}, which improve tabular reasoning via additional table-centric pre-training or instruction tuning.
(4) \textbf{Intermediate-Representation Methods}: \textit{GraphOTTER}~\cite{li-etal-2025-graphotter}, which represents tables as triples, and \textit{ST-Raptor}~\cite{tang2025straptorllmpoweredsemistructuredtable}, a tree-based baseline. 
 
\paragraph{Metrics.} Rigid string-matching metrics (e.g. Exact Match) often fail to capture semantic equivalence in generative tasks. To ensure robust evaluation, we employ LLMs as evaluators. Following recent methodologies validating the reliability of LLM-as-a-Judge frameworks~\cite{zheng2023judgingllmasajudgemtbenchchatbot, liu-etal-2023-g, dubois2025lengthcontrolledalpacaevalsimpleway}, we utilize GPT-5 as a binary judge: given a prediction and the gold answer, it outputs \texttt{CORRECT}/\texttt{INCORRECT}. We parse the label to compute Accuracy (Acc).

\paragraph{Implementations.} For fairness, we use \textbf{DeepSeek-V3-250324} as the backbone for all training-free methods, including our method (AdaSTR and DuTR), (\textit{E5}, \textit{EEDP}), and (\textit{GraphOTTER}, \textit{ST-Raptor}). Details are in Appendix~\ref{appendix:implementation_details}.

\begin{table}[t]
\centering
\resizebox{\linewidth}{!}{%
\begin{tabular}{l ccc}
\toprule
\textbf{Model} & \textbf{AIT-QA} & \textbf{SSTQA} & \textbf{HiTab} \\
\midrule
\multicolumn{4}{l}{\textit{\textbf{(1) Foundation Models}}} \\
DeepSeek-V3 & \cellcolor{bg-red}78.5 & \cellcolor{bg-red}63.2 & \cellcolor{bg-lgreen}82.0 \\
GPT-4o & \cellcolor{bg-red}80.6 & \cellcolor{bg-red}66.4 & \cellcolor{bg-yellow}78.6 \\
o3 & \cellcolor{bg-lgreen}89.1 & \cellcolor{bg-mgreen}78.2 & \cellcolor{bg-lgreen}85.3 \\
\midrule
\multicolumn{4}{l}{\textit{\textbf{(2) Prompting/Tool-augmented Methods}}} \\
E5 & \cellcolor{bg-lgreen}87.1 & \cellcolor{bg-yellow}70.2 & \cellcolor{bg-lgreen}85.1 \\
EEDP & \cellcolor{bg-lgreen}85.6 & \cellcolor{bg-lgreen}76.8 & \cellcolor{bg-yellow}79.2 \\
\midrule
\multicolumn{4}{l}{\textit{\textbf{(3) Table-Specific Adapted Models}}} \\
TableLlama$^\dagger$ & - & \cellcolor{bg-red}40.4 & \cellcolor{bg-red}64.7 \\
TableGPT2-72B$^\dagger$ & - & - & \cellcolor{bg-yellow}75.6 \\
\midrule
\multicolumn{4}{l}{\textit{\textbf{(4) Intermediate-Representation Methods}}} \\ 
GraphOTTER & \cellcolor{bg-mgreen}\underline{90.4} & \cellcolor{bg-yellow}71.5 & \cellcolor{bg-mgreen}88.8 \\
ST-Raptor & \cellcolor{bg-red}62.7 & \cellcolor{bg-yellow}71.1 & \cellcolor{bg-red}49.0 \\
\midrule
\multicolumn{4}{l}{\textit{\textbf{(5) ASTRA}}} \\
Textual Reasoning & \cellcolor{bg-lgreen}86.1 & \cellcolor{bg-mgreen}\underline{79.8} & \cellcolor{bg-lgreen}82.2 \\
Symbolic Reasoning & \cellcolor{bg-lgreen}87.3 & \cellcolor{bg-lgreen}75.3 & \cellcolor{bg-mgreen}\underline{89.3} \\
\textbf{Adaptive Selection} & \cellcolor{bg-mgreen}\textbf{91.6} & \cellcolor{bg-mgreen}\textbf{81.9} & \cellcolor{bg-mgreen}\textbf{90.1} \\
\midrule
\textit{Oracle} & \cellcolor{bg-dgreen}93.5 & \cellcolor{bg-dgreen}86.1 & \cellcolor{bg-dgreen}94.1 \\
\bottomrule
\end{tabular}%
}
\vspace{-1mm}
\caption{Accuracy (\%) on AIT-QA, SSTQA, and HiTab. \textbf{Bold} denotes best, \underline{underlined} denotes second best. $\dagger$ indicates results cited from prior literature. \textit{Oracle} represents the theoretical upper bound via ideal selection between Textual and Symbolic reasoning.}
\label{tab:main_results}
\vspace{-5mm}
\end{table}

\subsection{Main Results}

Table~\ref{tab:main_results} demonstrates \textbf{ASTRA}'s strong performance and trade-offs between the two reasoning modes. On SSTQA, rich in semantically intensive questions, Textual Reasoning ($79.8\%$) outperforms Symbolic Reasoning ($75.3\%$) by effectively capturing semantic nuances. Conversely, on HiTab, which features numerical aggregation, Symbolic Reasoning ($89.3\%$) surpasses Textual Reasoning ($82.2\%$), validating its computational accuracy. Notably, \textbf{ASTRA} ($90.1\%$ on HiTab) outperforms \textbf{o3} ($85.3\%$), validating the necessity of structural guidance. In contrast, rule-based \textbf{ST-Raptor} shows limited generalization on AIT-QA ($62.7\%$) and HiTab ($49.0\%$), exposing its fragility across diverse table layouts.

\subsection{Ablation Studies}
\label{sec:ablation_studies}

We conducted ablation studies on \textbf{SSTQA} to examine the impact of distinct components on tree reconstruction and reasoning efficacy.
\paragraph{Analysis of AdaSTR.}
Table~\ref{tab:ablation_tree_construction} presents the ablation results for the tree construction phase. The full \textbf{AdaSTR} pipeline achieves the highest performance across all metrics, demonstrating the synergy between the Evaluator-Guided Loop and the synthesis strategies.
\textbf{(1) Impact of Evaluator-Guided Loop:} Removing the evaluator reduces Average Coverage ($0.929 \to 0.745$), proving the loop effectively rectifies generation errors. \textbf{(2) Impact of Synthesis Strategies:} Disabling the adaptive synthesis strategies (i.e., defaulting to static DSP) leads to a substantial drop in the \textit{Minimum} Coverage Rate ($0.153$). This underscores that SRE and PSS are vital for handling large-scale complex tables where basic prompting fails.

\noindent\textbf{Remarks on Metric Absolute Values.} Results for Structural Accuracy and Coverage are partially influenced by artifacts in dataset annotation and the inherent effects of information filtering, rather than solely model generation errors. Detailed analysis on evaluation scores is provided in Appendix \ref{appendix:mode_selection}.

\begin{table}[t]
\centering
\resizebox{\linewidth}{!}{%
\begin{tabular}{lcccc}
\toprule
\multirow{2}{*}{\shortstack[l]{\textbf{Tree Construction} \\ \textit{\footnotesize{(DeepSeek-V3)}}}} & \multicolumn{2}{c}{\textbf{Coverage Rate}} & \multicolumn{2}{c}{\textbf{Struct Acc}} \\
\cmidrule(lr){2-3} \cmidrule(lr){4-5}
 & \textbf{Avg} & \textbf{Min} & \textbf{Avg} & \textbf{Min} \\
\midrule
\textbf{AdaSTR} & \textbf{0.929} & \textbf{0.738} & \textbf{0.792} & \textbf{0.654} \\
\quad w/o Evaluator-Guided & 0.745 & \underline{0.473} & \underline{0.698} & 0.143 \\
\quad w/o Synthesis Strategies & \underline{0.795} & 0.153 & 0.643 & \underline{0.457} \\
\bottomrule
\end{tabular}%
}
\caption{Ablation of \textbf{AdaSTR}. We report Average (Avg) and Minimum (Min) Coverage Rate and Structural Accuracy to evaluate robustness across diverse tables.}
\label{tab:ablation_tree_construction}
\vspace{-4mm}
\end{table}

\begin{table}[t]
\centering
\resizebox{\linewidth}{!}{%
\begin{tabular}{lcc}
\toprule
\textbf{Configuration} & \textbf{Acc (\%)} & \textbf{$\Delta$} \\
\midrule
\textbf{Adaptive Tree Navigation} & \textbf{79.84} & - \\
\quad w/o Embedding Model & 71.34 & \textcolor{red}{-8.50} \\
\quad w/o Dynamic Switching & - & - \\
\quad \quad \textit{Force Root-to-Leaf} & 77.23 & \textcolor{red}{-2.61} \\
\quad \quad \textit{Force Leaf-to-Root} & 75.65 & \textcolor{red}{-4.19} \\
\midrule
\textbf{Symbolic Tree Manipulation} & \textbf{75.26} & - \\
\quad w/o Code Examples & 70.42 & \textcolor{red}{-4.84} \\
\quad w/o Structural Skeleton & 71.73 & \textcolor{red}{-3.53} \\
\quad w/o Self-Correction Loop & 73.69 & \textcolor{red}{-1.57} \\
\midrule
\multicolumn{3}{l}{\textit{\textbf{Impact of Data Representation}}} \\
\quad \textit{Textual Serialization (Direct Prompting)} & 63.20 & - \\
\quad \textit{Semantic Tree (Direct Prompting)} & \textbf{70.55} & \textcolor{good}{+7.35} \\
\bottomrule
\end{tabular}%
}
\vspace{-1mm}
\caption{Ablation analysis of \textbf{DuTR (DeepSeek-V3)}. We report Accuracy (\%) and performance change ($\Delta$) vs. full configuration per mode. Bottom evaluates representational benefits of the tree structure.}
\label{tab:ablation_treeqa_deepseek}
\vspace{-3mm}
\end{table}

\paragraph{Analysis of DuTR.}
Table~\ref{tab:ablation_treeqa_deepseek} isolates the contributions of specific sub-modules:

\noindent\textbf{(1) Textual Reasoning Ablation:} Removing the path guide drops performance to $71.34\%$ ($\Delta -8.50\%$), confirming that semantic navigation provides essential global guidance. Furthermore, our dynamic variant ($79.84\%$) outperforms fixed strategies (max $77.23\%$), validating the necessity of query-adaptive traversal.

\noindent\textbf{(2) Symbolic Reasoning Ablation:} Omitting few-shot examples triggers the sharpest decline ($70.42\%$), followed by skeleton removal ($71.73\%$), highlighting format guidance and context filtering as critical drivers for effective program generation. 

\noindent\textbf{(3) Intrinsic Advantage of Data Representation.} To isolate representation gains, we compared static \textbf{Semantic Tree} inputs against \textbf{Textual Serialization}. Table~\ref{tab:ablation_treeqa_deepseek} confirms that even without advanced reasoning, the Static Tree ($70.55\%$) outperforms the Raw Table baseline ($63.20\%$), validating the intrinsic benefit of hierarchical serialization.

\subsection{Efficiency Analysis}

\begin{table}[t]
\centering
\resizebox{\linewidth}{!}{%
\begin{tabular}{l cc cc cc}
\toprule
\multirow{2}{*}{\textbf{Method}} & \multicolumn{2}{c}{\textbf{AIT-QA}} & \multicolumn{2}{c}{\textbf{HiTab}} & \multicolumn{2}{c}{\textbf{SSTQA}} \\
\cmidrule(lr){2-3} \cmidrule(lr){4-5} \cmidrule(lr){6-7}
 & \textbf{Tree} & \textbf{QA} & \textbf{Tree} & \textbf{QA} & \textbf{Tree} & \textbf{QA} \\
\midrule
ST-Raptor & 55.73 & 31.18 & 139.61 & 26.61 & 233.19 & 41.36 \\
GraphOTTER & - & 19.54 & - & 22.29 & - & 21.83 \\
\textbf{Ours} & \textbf{29.18} & \textbf{7.80} & \textbf{43.42} & \textbf{6.19} & \textbf{93.71} & \textbf{11.62} \\
\bottomrule
\end{tabular}%
}
\vspace{-1mm}
\caption{Comparison of time efficiency (in seconds) across different datasets. We record the latency for both the Tree Construction phase (Tree) and the Question Answering phase (QA). ``-'' denotes the method does not involve a tree construction phase.}
\label{tab:time_efficiency}
\vspace{-5mm}
\end{table}

Table~\ref{tab:time_efficiency} decomposes latency into offline Tree Construction and online QA Inference. By explicitly decoupling these stages into a ``write-once, read-many'' paradigm, our architecture ensures that the multi-stage structural parsing does not bottleneck online responsiveness. Our method ensures superior online efficiency and faster construction than ST-Raptor via PSS mode, avoiding costly element-wise recursion. Furthermore, empirical statistics indicate that the Evaluator-Guided Refinement Loop is triggered in only $\sim$7\% of cases, meaning the vast majority of trees are generated in a highly efficient single pass without iterative LLM overhead. While GraphOTTER builds faster, its high inference latency proves inefficient for multi-turn sessions. Amortized analysis ($T_{\text{avg}} = T_{\text{tree}}/N + T_{\text{qa}}$) confirms our method surpasses GraphOTTER when query count $N \ge 3$ (AIT-QA/HiTab) or $N \ge 10$ (SSTQA), offering an optimal accuracy-efficiency trade-off \textbf{for real-world, latency-sensitive applications}.



\subsection{Diagnostic Analysis}

\begin{figure}[t] 
    \centering 
    \includegraphics[width=1\linewidth]{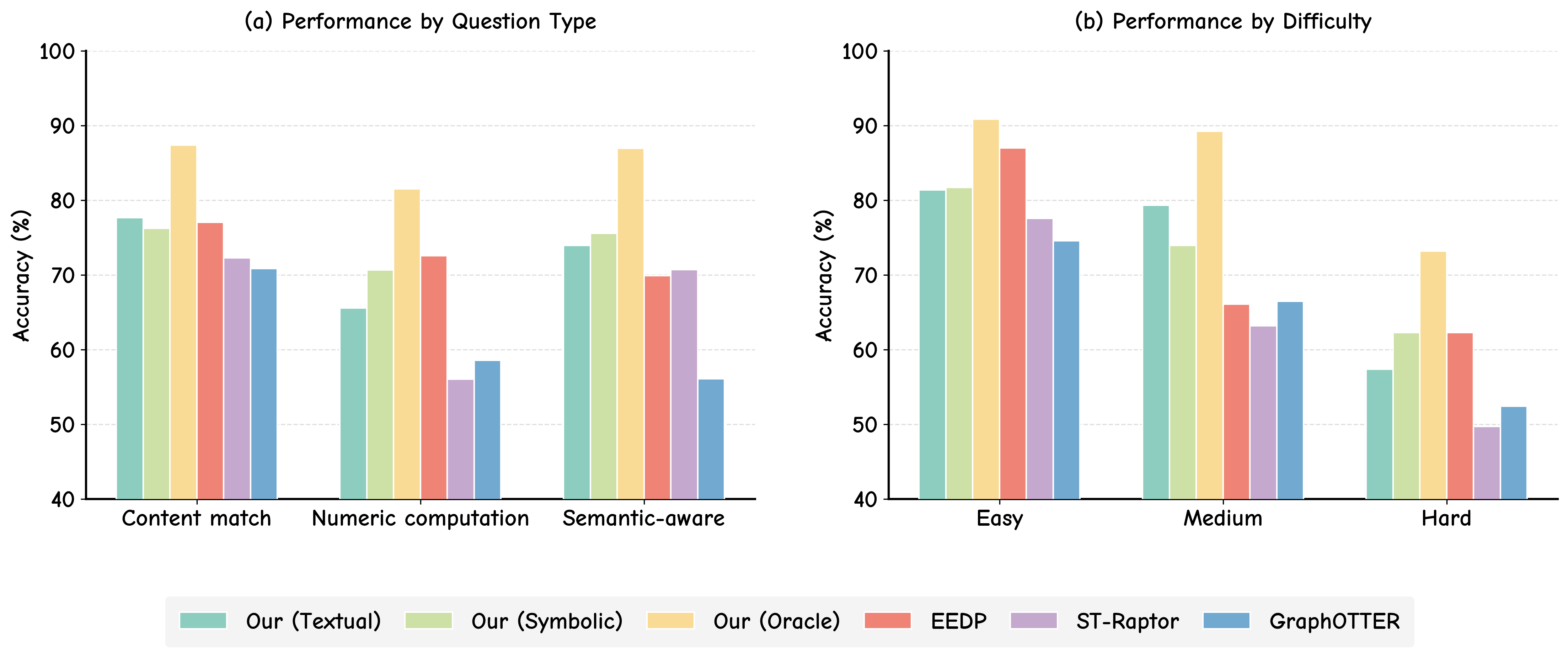} 
    \caption{Performance breakdown by question type and difficulty for DuTR and baselines.
} 
    \label{fig:diagnostic_analysis} 
    \vspace{-4mm}  
\end{figure}

We conduct a fine-grained analysis on SSTQA. 
(1) \textbf{Complementarity of Reasoning Modes.} Figure~\ref{fig:diagnostic_analysis} (a) confirms the synergy of Dual-Mode design. \textit{Textual Reasoning} excels in \textbf{Semantic} tasks, while \textit{Symbolic Reasoning} dominates \textbf{Numeric Computation}. Conversely, GraphOTTER struggles in semantic categories due to structural and semantic loss, a limitation resolved by our Semantic Tree.
(2) \textbf{Resilience to Query Complexity.} Figure~\ref{fig:diagnostic_analysis} (b) highlights robustness against increasing difficulty. While all models degrade on \textit{Hard} queries, our method maintains stability via synergistic textual-symbolic reasoning.
\begin{figure}[t]
    \centering
    \scriptsize
    \begin{tcolorbox}[
        colback=white, 
        colframe=black!70, 
        boxrule=0.8pt, 
        arc=2pt,
        width=\linewidth,
        title=\textbf{Case Study: Enumeration on Complex Hierarchy}
    ]
        \textbf{Query:} \textit{"How many items are listed in Fixed Expenses?"} \\
        \textbf{Gold Answer:} \textcolor{c_green}{\textbf{9}}
        
        \vspace{1mm}
        \fcolorbox{black!20}{bg_gray}{
        \parbox{0.95\linewidth}{
            \scriptsize \textbf{Table Snippet (Semantic Tree Representation):} \\
            \texttt{Manufacturing Overhead Budget} \\
            \texttt{\ \ $\vdash$ Variable Overhead \textcolor{gray}{// Sibling Branch}} \\
            \texttt{\ \ $\vert$\ \ $\vdash$ Indirect Labor ................ ...} \\
            \texttt{\ \ $\vert$\ \ $\llcorner$ ... (4 items omitted) ...} \\
            \texttt{\ \ $\llcorner$ Fixed Expenses \textcolor{c_blue}{// Target Branch}} \\
            \texttt{\ \ \ \ \ \ $\vdash$ Quarter 1} \\
            \texttt{\ \ \ \ \ \ $\vert$\ \ $\vdash$ Repair ........................ 2,500} \\
            \texttt{\ \ \ \ \ \ $\vert$\ \ $\vdash$ ... (8 standard items) ...} \\
            \texttt{\ \ \ \ \ \ $\vert$\ \ $\llcorner$ Cash Output ................. 16,635} \\
            \texttt{\ \ \ \ \ \ $\llcorner$ ... (Quarter 2-4 omitted) ...}
        }}
        
        \vspace{1.5mm}
        \hrule
        \vspace{1.5mm}

        \textbf{\textcolor{c_red}{Baselines (Failure Analysis)}}
        \begin{itemize}[leftmargin=*, nosep, itemsep=2pt]
            \item \textbf{GraphOTTER (Output: 5):} \textit{Representation Gap.} \\
            Retrieves only standard items (e.g., "Repair") but misses structurally bound rows (e.g., "Total", "Less").
            
            \item \textbf{ST-Raptor (Output: 1):} \textit{Structural Neglect.} \\
            Overlooks the implicit parent-child hierarchy, treating the header ``Fixed Expenses" as a single data point rather than a container.
            
            \item \textbf{EEDP (Output: 5):} \textit{Reasoning Opacity.} \\
            Relies on textual commonsense to identify ``expenses" during CoT generation, ignoring structural aggregations (e.g., ``Total") that don't look like standard items.
        \end{itemize}

        \vspace{1.5mm}
        \hrule
        \vspace{1.5mm}

        \textbf{\textcolor{c_green}{ASTRA (Ours)}}
        \begin{itemize}[leftmargin=*, nosep]
            \item \textbf{Logic (Symbolic Tree Manipulation):} \\
            \texttt{\quad \textcolor{c_purple}{target} = tree[\textquotesingle Manuf. Budget\textquotesingle][\textquotesingle Fixed Expenses\textquotesingle][\textquotesingle Q1\textquotesingle]} \\
            \texttt{\quad \textcolor{c_purple}{Answer} = len(\textcolor{c_purple}{target}.keys())}
            
            \item \textbf{Result:} \textbf{Correct (9)}. The dictionary structure strictly enforces parent-child containment, ensuring all irregular items are counted.
        \end{itemize}
    \end{tcolorbox}
    \vspace{-3mm}
    \caption{A representative case study involving sibling branches and irregular children. \textbf{ASTRA} utilizes precise dictionary traversal to correctly enumerate all items.}
    \label{fig:case_study_box}
    \vspace{-5mm}
\end{figure}

\subsection{Case Study}
\label{sec:case_study}
To intuitively demonstrate how ASTRA addresses the serialization bottlenecks, we present a case on \textbf{Structural Enumeration} in Figure~\ref{fig:case_study_box}. The query requires enumerating all children under the parent category ``Fixed Expenses".
GraphOTTER fails due to the \textbf{Representation Gap}; its triple-based format isolates cells as discrete entities, making it unable to capture the structural dependency of aggregation rows. ST-Raptor exhibits \textbf{Structural Neglect}, erroneously regarding the parent header "Fixed Expenses" as an atomic data point. EEDP suffers from \textbf{Reasoning Opacity}; by relying on purely textual inference, it introduces \textit{semantic bias}—prioritizing the model's parametric internal knowledge over the table's explicit structural constraints—thereby hallucinating filters that exclude valid but irregular items. In contrast, ASTRA leverages its Semantic Tree representation to preserve explicit parent-child relationships. By navigating the tree structure, it successfully retrieves the complete subtree, demonstrating the critical need for \textbf{Explicit Hierarchy} and \textbf{Semantic Context} when handling complex nested headers.

\section{Conclusion}
We present ASTRA, a training-free method for semantic tree reconstruction that introduces a Text-Symbolic Reasoning paradigm to enhance tabular QA. By restoring explicit hierarchy, our approach achieves SOTA performance, outperforming advanced reasoning models like OpenAI o3. Notably, we find that the Semantic Tree representation alone—even without specialized reasoning mechanisms—yields superior performance compared to textual serializations. These findings offer a critical insight: an explicit hierarchy enriched with semantic context is critical for unlocking the full reasoning potential of LLMs.


\section*{Limitations}
While ASTRA achieves state-of-the-art performance on complex table benchmarks, there remain avenues for future exploration tied to the inherent challenges of the task setting. First, regarding efficiency trade-offs: Our Semantic Tree construction is specifically optimized to disentangle the intricate dependencies in complex, hierarchical tables. For extremely simple flat tables where explicit hierarchy is absent, this reconstruction process may incur a computational overhead compared to direct textual serialization. Second, ASTRA primarily relies on textual and structural parsing. However, real-world complex tables often utilize visual cues (e.g., background colors, bold fonts) to convey implicit semantic constraints. Incorporating multi-modal vision encoders to capture these stylistic features remains a promising direction. 

\section*{Ethics Statement}
In this paper, we introduce ASTRA, a framework for complex table question answering. Our experimental evaluation is conducted using publicly available and widely recognized benchmarks, including AIT-QA, HiTab, and SSTQA. These datasets are constructed from open-domain sources (e.g., Wikipedia, public financial reports) and, to the best of our knowledge, do not contain any personally identifiable information or offensive content. Furthermore, all Large Language Models employed in this work were utilized in strict adherence to their respective usage policies and safety guidelines. We believe that our work does not pose any significant ethical concerns.

\section*{Acknowledgments}
This work is funded by National Natural Science Foundation of China (NSFC62306276/NSFCU23B2055), New Generation Artificial Intelligence-National Science and Technology Major Project 2030 (2025ZD0122800), Yongjiang Talent Introduction Programme (2022A-238-G), and Fundamental Research Funds for the Central Universities (226-2023-00138). This work was supported by Ant Group.

\bibliography{custom}
\appendix

\section{Implementation Details}
\label{appendix:implementation_details}
All experiments were conducted on Ubuntu 20.04.1 LTS servers equipped with two NVIDIA A100 GPUs. We accessed large-scale open-source and proprietary models via respective APIs. For smaller models, specifically Qwen3-8B, Qwen2.5-7B, InternVL2.5-26B, and Llama-3.1-8B,  we deployed them locally on the servers.

\subsection{Implementation of Baselines}

All baseline methods were implemented using the official codebases provided by the original authors, adhering to the default hyperparameter settings unless otherwise noted. Specific adaptations are detailed below:

(1) \textbf{EEDP}~\cite{srivastava2025evaluatingllmsmathematicalreasoning}: Due to the absence of official source code, we reproduced the method relying strictly on the prompts documented in the original paper. 

(2) \textbf{Foundation Model}: We utilize standard \textbf{textual serialization} (e.g., HTML or Markdown) to represent tabular data, evaluating the models' intrinsic reasoning capabilities without structural augmentation. The detailed prompt templates used for these baselines are provided in Listing~\ref{lst:foundation_prompt}.


\subsection{Implementation of Our method}

\begin{algorithm}[h]
\caption{Adaptive Tree Navigation}
\label{alg:treeqa}
\begin{algorithmic}[1]
\REQUIRE Tree Structure $\mathcal{T}$, Question $Q$
\ENSURE Evidence Context $\mathcal{C}$, Answer $A$
\STATE Preprocess $\mathcal{T}$ (normalize values)
\STATE $d \leftarrow \Func{DetermineDir}(\mathcal{T}, Q)$
\IF{$d = \text{"Root2Leaf"}$}
    \STATE $(\mathcal{C}, A) \leftarrow \Func{Root2Leaf}(\mathcal{T}, Q)$
\ELSE
    \STATE $(\mathcal{C}, A) \leftarrow \Func{Leaf2Root}(\mathcal{T}, Q)$
\ENDIF
\RETURN $(\mathcal{C}, A)$
\end{algorithmic}
\end{algorithm}

\paragraph{AdaSTR.} We utilized \textit{DeepSeek-V3} for the tree construction phase. To ensure deterministic outputs, the temperature was set to $0$. For the validation loop, we set the acceptance thresholds for Information Coverage and Structural Integrity to $80\%$ and $70\%$, respectively, with a maximum of $3$ reconstruction attempts. If none of the attempts satisfies both thresholds within three trials, we select the candidate from the attempted reconstructions that achieves the highest average score across the two metrics. Upon completion, the constructed trees were cached in JSON format to facilitate efficient retrieval in downstream tasks. Specific details regarding the adaptive strategy selection policy are provided in Appendix~\ref{appendix:mode_selection}.

\paragraph{DuTR.} For the question-answering phase within our proposed framework (Algorithm~\ref{alg:treeqa}), we employed \textit{DeepSeek-V3} with the temperature set to $0.3$. In the textual reasoning stage (detailed in Algorithm~\ref{alg:root2leaf} and Algorithm~\ref{alg:leaf2root}), \textit{Multilingual-E5-Large}~\cite{wang2024multilinguale5textembeddings} served as the embedding model to support path retrieval and leaf ranking. Regarding retrieval hyperparameters, we set $k_2=5$ for the Root2Leaf strategy, while adopting a larger $k_1=50$ for Leaf2Root to ensure comprehensive coverage of candidate leaf nodes.

To determine the final answer from the candidate outputs (Textual vs. Symbolic), we employed a locally deployed lightweight model, \textbf{Qwen3-8B}, as the answer selector. This approach is highly efficient, introducing negligible computational overhead; statistical analysis shows an average execution time of only \textbf{0.42s} for AIT-QA, \textbf{0.51s} for HiTab, and \textbf{0.63s} for SSTQA per query. The prompt used for this adaptive selection is as follows:

\begin{tcolorbox}[
    colback=white,        
    colframe=black,       
    boxrule=0.8pt,        
    arc=0pt,              
    left=6pt, right=6pt, top=6pt, bottom=6pt, 
    fontupper=\small\ttfamily 
]
    
    You are a rigorous data verification assistant specializing in complex JSON tree-structured table data. Your task is to determine which candidate answer is correct based on the provided table facts.
    
    \vspace{0.5em}
    \textbf{\#\# Table:}
    
    \{table\}
    
    \textbf{\#\# Question:}
    
    \{question\}
    
    \textbf{\#\# Candidate Answers:}
    
    \textbf{Answer A}: \{answerA\}
    
    \textbf{Answer B}: \{answerB\}
    
    \vspace{0.5em}
    \textbf{Analysis Steps:}
    
    1. Locate relevant data within the table.
    
    2. Compare the consistency of both answers against the table data.
    
    \vspace{0.5em}
    \textbf{Output Requirements:}
    
    Do not output any explanations, punctuation, or analysis processes. Strictly output ONLY a single character: "A" or "B".
    
    \vspace{0.5em}
    \textbf{The Correct Answer:}
\end{tcolorbox}

We also conduct preliminary explorations on improving the selector; details are provided in Appendix~\ref{app:selector_optimization}.

\subsection{Implementation of Evaluation Metrics}

Following recent best practices validating the reliability of automated evaluation~\cite{zheng2023judgingllmasajudgemtbenchchatbot, liu-etal-2023-g}, we adopted an \textbf{LLM-as-a-Judge} paradigm to assess answer accuracy, utilizing \textit{GPT-5} with a temperature of $0.3$. To explicitly validate the reliability of this automated metric, we conducted a \textbf{manual verification} on a subset of 200 queries randomly sampled from the \textbf{SSTQA dataset}. Since our framework generates two candidate answers per query (derived from \textit{Textual} and \textit{Symbolic} reasoning modes, respectively), this validation set comprised a total of 400 generated answers. We manually reviewed these instances to establish ground-truth correctness. As shown in our analysis, the automated judge demonstrated an exceptionally high alignment with human decision-making, achieving an agreement rate of \textbf{98.25\%} ($393/400$). Notably, the discrepancies were minimal and balanced, comprising only 3 false positives (judged correct but actually incorrect) and 4 false negatives (judged incorrect but actually correct), indicating no significant bias in the evaluator. Given this near-perfect consistency, we proceeded with the automated evaluation for the full dataset. The exact prompt used for judgment is:

\begin{tcolorbox}[
    colback=white,        
    colframe=black,       
    boxrule=0.8pt,        
    arc=0pt,              
    left=6pt, right=6pt, top=6pt, bottom=6pt, 
    fontupper=\small\ttfamily 
]
    \textbf{Judge Prompt:}
    
    \vspace{0.5em}
    Question: \{\$question\}
    
    Correct Answer: \{\$ground\_truth\}
    
    Predicted Answer: \{\$prediction\}
    
    \vspace{0.5em}
    Please judge whether the predicted answer is correct. If the answer involves numerical values, slight numerical errors are allowed. Please answer only "Correct" or "Incorrect".
\end{tcolorbox}

\section{Baseline Descriptions}
\label{app:baseline_descriptions}

In this section, we provide detailed descriptions of the baseline methods used in our experiments:
\subsection{Prompting and Tool-augmented Methods}
\begin{itemize}
    \item \textbf{E5}~\cite{zhang-etal-2024-e5} is a zero-shot, code-augmented framework specifically designed for hierarchical table analysis. It operates via a five-step pipeline: \textit{Explain} the table structure, \textit{Extract} information via code generation, \textit{Execute} the code to ensure accuracy, \textit{Exhibit} the results, and \textit{Extrapolate} to derive the final answer. It effectively addresses the challenges of complex table structures and implicit semantics without requiring fine-tuning.
    
    \item \textbf{EEDP}~\cite{srivastava2025evaluatingllmsmathematicalreasoning} is a Chain-of-Thought (CoT) prompting strategy tailored for reasoning over semi-structured financial documents. It guides the LLM through four logical steps: \textit{Elicit} domain knowledge, \textit{Extract} relevant evidence (rows/numbers), \textit{Decompose} the problem into atomic calculations, and \textit{Predict} the final answer.
\end{itemize}

\subsection{Table-Specific Adapted Models}
\begin{itemize}
    \item \textbf{TableLlama}~\cite{zhang2024tablellamaopenlargegeneralist} is an open-source LLM fine-tuned on TableInstruct, a large-scale dataset comprising diverse table tasks. It treats tables as serialized text sequences and utilizes LongLoRA to efficiently handle the long context windows typically required for processing large tables.
    
    \item \textbf{TableGPT2}~\cite{su2024tablegpt2largemultimodalmodel} is a 72B-parameter multimodal model that treats tables as a distinct modality. It integrates a dedicated table encoder with the LLM decoder and is pre-trained on massive table-text pairs, allowing it to capture schema-level structural information and perform robust numerical reasoning.
\end{itemize}

\subsection{Intermediate-Representation Methods}
\begin{itemize}
    \item \textbf{GraphOTTER}~\cite{li-etal-2025-graphotter} addresses complex layouts by converting tables into an undirected graph representation, where cells are nodes connected by spatial edges. It employs an LLM-driven agent to iteratively traverse this graph and perform reasoning, effectively handling merged cells and headers.
    
    \item \textbf{ST-Raptor}~\cite{tang2025straptorllmpoweredsemistructuredtable} models semi-structured tables as Hierarchical Orthogonal Trees (HO-Trees) to explicitly capture nesting relationships. It resolves queries by decomposing them into atomic tree operations (e.g., retrieval, aggregation) which are executed deterministically against the structured tree representation.
\end{itemize}

\section{Dataset Details}
\label{app:dataset_details}

\begin{table}[t]
    \centering
    \resizebox{\columnwidth}{!}{
    \begin{tabular}{l rr l}
        \toprule
        \textbf{Dataset} & \textbf{\# Tables} & \textbf{\# QAs} & \textbf{Domain} \\
        \midrule
        HiTab & 538 & 1,584 & Open domain \\
        AIT-QA & 80 & 367 & Airline industry \\
        SSTQA & 102 & 764 & Open domain \\
        \bottomrule
    \end{tabular}
    }
        \caption{\textbf{Dataset Statistics.} Details of the test sets used in our experiments. SSTQA specifically targets real-world semi-structured tables.}
    \label{tab:dataset_details}
\end{table}

We evaluate on three public table QA benchmarks (Table~\ref{tab:dataset_details}).
Below we summarize their characteristics.

\paragraph{HiTab (test split).}
HiTab is an open-domain benchmark focusing on \emph{hierarchical tables}, where multi-level row/column headers and implicit structure make numerical reasoning and indexing challenging.
The tables are collected from real statistical reports and Wikipedia, and the QA pairs are constructed by rewriting analyst-authored descriptive sentences into questions, which helps preserve real-world analytical intent.
The released annotations include fine-grained entity/quantity alignments; each QA instance additionally records its aggregation operator and may provide an explicit answer formula.
In our setting, we report results on the HiTab \textbf{test set} (538 tables / 1,584 QA pairs as in Table~\ref{tab:dataset_details}).

\begin{algorithm}[h]
\caption{Leaf-to-Root Reasoning}
\label{alg:leaf2root}
\begin{algorithmic}[1]
\REQUIRE Tree Structure $\mathcal{T}$, Question $Q$
\ENSURE Evidence $\mathcal{C}$, Answer $A$

\STATE $\mathcal{L} \leftarrow \Func{GetAllLeaves}(\mathcal{T})$
\IF{using embedding}
    \STATE $\mathcal{L} \leftarrow \Func{EmbSort}(\mathcal{L}, Q)$ \COMMENT{Rank leaf nodes by relevance}
\ENDIF

\STATE $\mathcal{L}_{rel} \leftarrow \LLM{filter}(\mathcal{L}, Q)$ \COMMENT{Filter irrelevant leaves}
\STATE $k \leftarrow 0, K_{max} \leftarrow 5$

\WHILE{$k \le K_{max}$}
    \STATE $\mathcal{C}_{k} \leftarrow \emptyset$
    \FOR{$(p_{leaf}, v) \in \mathcal{L}_{rel}$}
        \STATE $p_{sub} \leftarrow p_{leaf}[:|p_{leaf}| - k]$ \COMMENT{Prune path upward by depth $k$}
        \STATE $v_{sub} \leftarrow \Func{GetData}(\mathcal{T}, p_{sub})$
        \STATE $\mathcal{C}_{k} \leftarrow \mathcal{C}_{k} \cup \{(p_{sub}, v_{sub})\}$
    \ENDFOR
    
    \STATE $\mathcal{C} \leftarrow \Func{Merge}(\mathcal{C}_{k})$ \COMMENT{Merge overlapping paths}
    
    \STATE $(ready, A) \leftarrow \LLM{check}(Q, \mathcal{C})$
    \IF{$ready$}
        \RETURN $(\mathcal{C}, A)$
    \ENDIF
    
    \STATE $k \leftarrow k + 1$ \COMMENT{Increase depth to expand context}
\ENDWHILE

\STATE $A \leftarrow \LLM{gen}(Q, \mathcal{C})$ \COMMENT{Fallback generation}
\RETURN $(\mathcal{C}, A)$
\end{algorithmic}
\end{algorithm}

\paragraph{AIT-QA.}
AIT-QA is a domain-specific benchmark built from tables in public U.S. SEC filings of major airline companies (fiscal years 2017--2019).
Compared with Wikipedia-style flat tables, AIT-QA tables often have more complex layouts, including hierarchical headers and domain-specific terminology.
The dataset also provides metadata/annotations indicating whether a question requires hierarchical header understanding, involves domain terminology, or is a paraphrase.
In our experiments we follow the common evaluation subset reported in Table~\ref{tab:dataset_details} (80 tables / 367 QA pairs).

\paragraph{SSTQA.}
SSTQA targets \emph{real-world semi-structured tables} (e.g., spreadsheet-like layouts) that contain irregular structures such as merged cells, multi-row/column headers, nested regions (subtables), and flexible layouts.
It contains 764 questions over 102 curated tables, selected from a larger pool of real-world tables to ensure diverse structures and broad scenario coverage (e.g., administration, finance, HR, schedules, etc.).
This benchmark is designed to stress layout-aware reasoning beyond the assumptions of fully structured (relational) tables.

\begin{algorithm}[h]
\caption{Root-to-Leaf Reasoning}
\label{alg:root2leaf}
\begin{algorithmic}[1]
\REQUIRE Tree Structure $\mathcal{T}$, Question $Q$
\ENSURE Evidence $\mathcal{C}$, Answer $A$
\STATE $S \leftarrow [(\text{root}, \mathcal{T}, [\text{root}])]$ \COMMENT{Init Stack}
\STATE $\mathcal{C} \leftarrow \emptyset$ \COMMENT{Accumulated Evidence Context}
\STATE $\mathcal{G} \leftarrow \emptyset$ \COMMENT{Global Semantic Guidance}

\IF{using embedding}
    \STATE $\mathcal{P}_{topk} \leftarrow \Func{EmbRank}(\mathcal{T}, Q)$ \COMMENT{Retrieve top-$k$ relevant paths}
    \STATE $\mathcal{G} \leftarrow \Func{ExtractRoots}(\mathcal{P}_{topk})$
\ENDIF

\WHILE{$S \neq \emptyset$}
    \STATE $(n, v, p_{curr}) \leftarrow \Func{Pop}(S)$
    
    \IF{$\Func{IsLeaf}(n)$}
        \STATE $\mathcal{C} \leftarrow \mathcal{C} \cup \{(p_{curr}, v)\}$
        
        \STATE $(ready, A_{temp}) \leftarrow \LLM{check}(Q, \mathcal{C})$
        \IF{$ready$}
            \RETURN $(\mathcal{C}, A_{temp})$ 
        \ENDIF
    \ELSE
        \STATE $\mathcal{N}_{child} \leftarrow \Func{GetChildren}(v)$
        \STATE $\mathcal{N}_{sel} \leftarrow \LLM{select}(n, \mathcal{N}_{child}, Q, \mathcal{G})$
        \FOR{$c \in \mathcal{N}_{sel}$}
            \STATE $\Func{Push}(S, (c, \mathcal{N}_{child}[c], p_{curr} \cup \{c\}))$
        \ENDFOR
    \ENDIF
\ENDWHILE

\STATE $A \leftarrow \LLM{gen}(\mathcal{C}, Q)$
\RETURN $(\mathcal{C}, A)$
\end{algorithmic}
\end{algorithm}

\section{Detailed Comparison between ST-Raptor and AdaSTR} 
\label{appendix:difference}

This appendix elucidates the key distinctions between ST-Raptor~\cite{tang2025straptorllmpoweredsemistructuredtable} and our proposed AdaSTR framework through a concrete case study. ST-Raptor employs a rule-based, open-loop approach to construct hierarchical trees (HO-Trees) primarily reliant on physical layout heuristics, such as cell merging annotations and positional alignment. While effective for simple layouts, this method introduces vulnerabilities in complex tables where semantic dependencies may not align with a predefined rule, leading to local bias and error propagation from initial detections.

\begin{figure*}[h]
\centering
\includegraphics[width=0.8\textwidth]{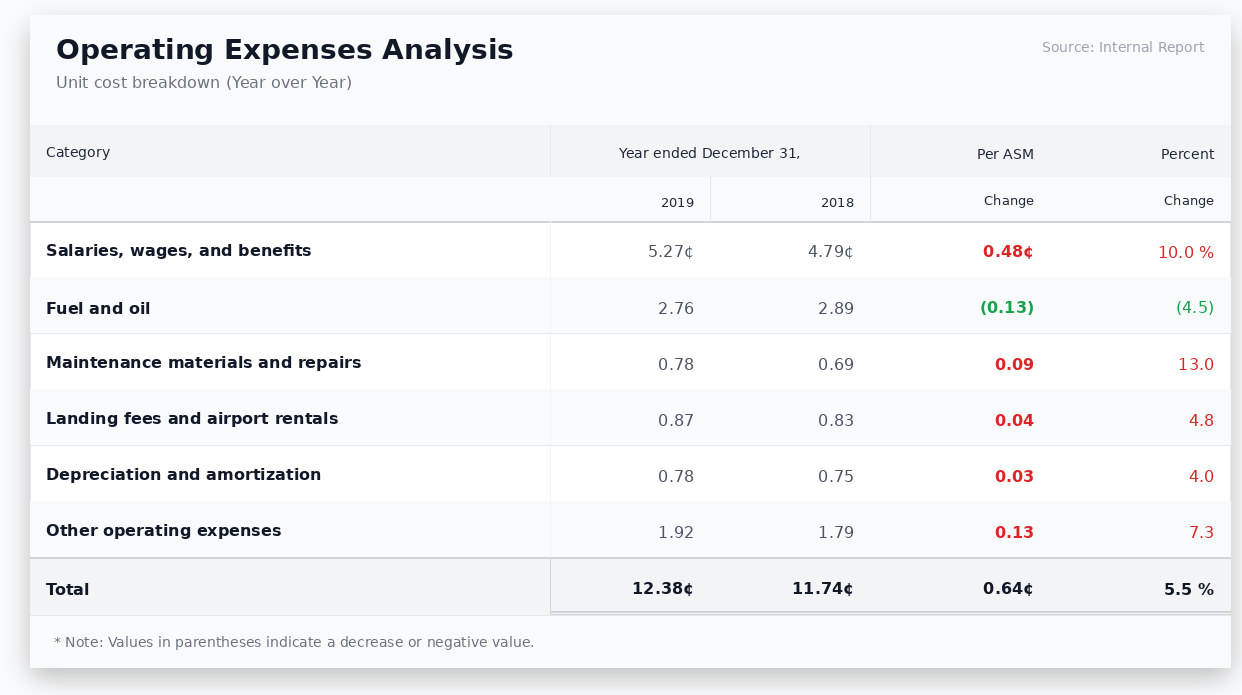} 
\caption{Visualization of the Operating Expenses Analysis table.}
\label{fig:financial_table}
\end{figure*}

In contrast, AdaSTR reframes table parsing as a closed-loop "Semantic Reconstruction" task, leveraging LLMs' global context-awareness to infer latent logical subordination directly from content. This results in a more robust semantic tree where nodes represent subjects and attributes, edges denote logical relationships (e.g., subordination, qualifications), and leaves hold data values. The adaptive mechanism in AdaSTR further ensures scalability by dynamically selecting construction strategies based on table characteristics, with feedback loops for quality assurance.
To illustrate, we use a real-world financial table: an "Operating Expenses Analysis" breakdown comparing unit costs year-over-year (2019 vs. 2018), including per ASM changes and percent changes. The table features merged headers (e.g., "Year ended December 31," spanning 2019 and 2018 columns) and irregular alignments, making it representative of complex, semi-structured data in finance domains.
\paragraph{Table description.}
The input is an HTML-rendered table (visualized in Figure~\ref{fig:financial_table}). Key structural elements include: A top-level header "Year ended December 31," merged across two sub-columns (2019 and 2018).
Separate columns for "Per ASM Change" and "Percent Change."
Rows for categories like "Salaries, wages, and benefits," with values in cents (¢) or plain numbers, and parentheses indicating decreases.

\paragraph{ST-Raptor: geometry-driven HO-Tree leads to semantic detachment.}
Listing~\ref{lst:straptor_tree} shows the output of ST-Raptor.
We highlight several concrete failure modes:
(i) \textbf{Unanchored header node:} a placeholder node (\texttt{"None"}) is created for \{2019, 2018, change\} without binding them to the body cells, indicating the header hierarchy is not grounded to the data region.
(ii) \textbf{Positional keys instead of semantic attributes:} for most rows, columns are indexed by \texttt{"0","1","2","3"}, which removes column semantics and prevents deterministic querying.
(iii) \textbf{Value-as-key corruption:} for \textit{Depreciation and amortization}, numerical values (\texttt{0.78, 0.75, 0.03, 4.0}) are mistakenly promoted to keys with empty values, suggesting that local alignment cues override global header constraints once header-cell mapping fails.
(iv) \textbf{Schema inconsistency:} the \textit{Total} row uses \texttt{value1..value4}, a schema incompatible with other rows, reflecting the lack of a global consistency check.
These errors are consistent with ST-Raptor's open-loop, rule-based construction that assumes physical layout reliably reflects logical hierarchy; once header detection is imperfect, the error cascades irreversibly.

\paragraph{AdaSTR: semantic reconstruction leads to coherent structural alignment.}
Listing~\ref{lst:adastr_tree} shows AdaSTR's reconstructed semantic tree.
AdaSTR represents each row subject (\textit{Category}) as a node, and attaches fully-qualified attributes derived from the header path, e.g.,
\texttt{Year ended December 31, - 2019} and \texttt{Year ended December 31, - 2018}.
Crucially, repeated headers are disambiguated by parent context (\texttt{Per ASM - change} vs. \texttt{Percent - change}).
All categories share a consistent attribute set, making the representation directly consumable by symbolic querying and program synthesis.

\paragraph{Takeaway.}
This case illustrates that AdaSTR better satisfies the four desiderata for complex table serialization:
(1) \textbf{Explicit hierarchy \& semantic context} via header-path attributes;
(2) \textbf{Representation Alignment} by enabling natural-language-like statements such as ``(For) Fuel and oil, Year ended December 31 2019 (is) 2.76; ...'';
(3) \textbf{Schema flexibility} since attributes are attached per subject node without forcing a rigid relational schema;
(4) \textbf{Symbolic compatibility} because the tree can be deterministically converted to key-value records or executable queries.

\begin{figure*}[t]
    \centering
    \begin{minipage}[t]{0.48\textwidth}
        \centering
        \captionof{listing}{ST-Raptor output (Baseline). Note the semantic loss in headers ("None") and structural corruption in the "Depreciation" row (lines 28-33).}
        \label{lst:straptor_tree}
        \begin{minted}[
            fontsize=\scriptsize, % 使用更小的字体以适应宽度
            linenos,              % 显示行号，方便引用
            frame=lines,          % 添加上下边框
            framesep=2mm,
            baselinestretch=1.0,  % 行间距
            breaklines,            % 自动换行
            xleftmargin=15pt   % <==== 新增这一行：将整个代码块向右平移 15pt
        ]{json}
{
    "table": {
        "None": { 
        "2019": "", "2018": "", "change": "" 
        },
        "Salaries, wages, and benefits": {
            "0": "5.27¢", "1": "4.79¢", 
            "2": "0.48¢", "3": "10.0 %"
        },
        "Fuel and oil": {
            "0": "2.76", "1": "2.89", 
            "2": "(0.13)", "3": "(4.5)"
        },
        "Maintenance materials and repairs": {
            "0": "0.78", "1": "0.69", 
            "2": "0.09", "3": "13.0"
        },
        "Landing fees and airport rentals": {
            "0": "0.87", "1": "0.83", 
            "2": "0.04", "3": "4.8"
        },
        "Depreciation and amortization": {
            // CRITICAL FAILURE HERE
            "0.78": "", "0.75": "", 
            "0.03": "", "4.0": ""
        },
        "Other operating expenses": {
            "0": "1.92", "1": "1.79", 
            "2": "0.13", "3": "7.3"
        },
        "Total": {
            "value1": "12.38¢", "value2": "11.74¢",
            "value3": "0.64¢", "value4": "5.5 %"
        }
    }
}
        \end{minted}
    \end{minipage}
    \hfill 
    \begin{minipage}[t]{0.48\textwidth}
        \centering
        \captionof{listing}{AdaSTR output (Ours). Note the fully preserved semantic paths and correct value alignment across all rows.}
        \label{lst:adastr_tree}
        \begin{minted}[
            fontsize=\scriptsize,
            linenos,
            frame=lines,
            framesep=2mm,
            baselinestretch=1.0,
            breaklines,
            xleftmargin=15pt   % <==== 新增这一行：将整个代码块向右平移 15pt
        ]{json}
{
  "tree_table": {
    "Salaries, wages, and benefits": {
      "Year ended... - 2019": "5.27¢",
      "Year ended... - 2018": "4.79¢",
      "Per ASM - change": "0.48¢",
      "Percent - change": "10.0 %"
    },
    "Fuel and oil": {
      "Year ended... - 2019": "2.76",
      "Year ended... - 2018": "2.89",
      "Per ASM - change": "(0.13)",
      "Percent - change": "(4.5)"
    },
    "Maintenance materials...": {
      "Year ended... - 2019": "0.78",
      "Year ended... - 2018": "0.69",
      "Per ASM - change": "0.09",
      "Percent - change": "13.0"
    },
    "Landing fees...": {
      "Year ended... - 2019": "0.87",
      "Year ended... - 2018": "0.83",
      "Per ASM - change": "0.04",
      "Percent - change": "4.8"
    },
    "Depreciation and amortization": {
      // CORRECT ALIGNMENT
      "Year ended... - 2019": "0.78",
      "Year ended... - 2018": "0.75",
      "Per ASM - change": "0.03",
      "Percent - change": "4.0"
    },
    "Other operating expenses": {
      "Year ended... - 2019": "1.92",
      "Year ended... - 2018": "1.79",
      "Per ASM - change": "0.13",
      "Percent - change": "7.3"
    },
    "Total": {
      "Year ended... - 2019": "12.38¢",
      "Year ended... - 2018": "11.74¢",
      "Per ASM - change": "0.64¢",
      "Percent - change": "5.5 %"
    }
  }
}
        \end{minted}
    \end{minipage}
\end{figure*}

\section{Comprehensive Taxonomy of Reasoning Methods}
\label{appendix:reasoning_method}

Table question answering (TableQA) in the LLM era can be understood as a pipeline of two design choices:
(i) how the table is \emph{represented} to the model, (ii) how the model \emph{reasons} over that representation. 
Following recent TableQA literature~\cite{liu2023rethinkingtabulardataunderstanding,zhang-etal-2024-syntqa},
we summarize two major paradigms—\textbf{textual (end-to-end) reasoning} and \textbf{symbolic (program-aided) reasoning}—and
highlight \textbf{hybrid} variants that combine the strengths of both.

\vspace{0.5em}
\noindent\textbf{Notation.}
We denote a table as $T$ with cells $T[r,c]$, headers $\mathcal{H}$ (flat or multi-level),
and a natural language question $q$. A TableQA system outputs an answer $\hat{a}$ and optionally a rationale $\hat{y}$ or an executable
program $\hat{p}$ such that $\hat{a}=\texttt{Exec}(\hat{p},T)$.

\paragraph{1) Textual reasoning (End-to-End approaches).}
Textual reasoning treats TableQA as conditional generation:
\begin{equation}
(\hat{y},\hat{a}) = \texttt{LLM}\big(q \,\Vert\, \texttt{Serialize}(T)\big),
\end{equation}
where $\texttt{Serialize}(\cdot)$ linearizes the table into a token sequence (e.g., Markdown/CSV/TSV, row-wise templates,
or hierarchical header strings). The LLM produces a natural-language reasoning trace $\hat{y}$ (optional) and the final
answer $\hat{a}$ directly. In practice, this paradigm is often implemented via direct prompting, chain-of-thought (CoT),
self-consistency, or decomposition prompts~\cite{wei2023chainofthoughtpromptingelicitsreasoning,chen-2023-large}.
Recent work emphasizes that careful \emph{table provisioning}—selecting relevant rows/columns, augmenting metadata, and packing
contexts—can substantially improve end-to-end reasoning accuracy under context limits~\cite{sui-etal-2024-tap4llm}. More recently, \emph{Row-of-Thought} (RoT) prompting structures the generation as iterative \emph{row-wise traversals} with
intermediate reflection/refinement, effectively scaling reasoning length while keeping the model grounded in the table and
reducing table-content hallucinations in a training-free manner~\cite{zhang2025rotenhancingtablereasoning}.

\noindent\textbf{Strengths.}
Textual reasoning is semantically flexible: it can handle vague questions, implicit references, and multi-hop explanations
without requiring a fixed operator set. It is also easy to deploy (no executor/tooling) and can be robust to minor schema
variations because the model reasons over natural language rather than strict schemas.

\noindent\textbf{Failure modes.}
Despite strong semantic understanding, end-to-end generation may suffer from:
\begin{itemize}
  \item \textbf{Numerical and arithmetic errors}: LLMs may perform approximate calculation or lose track of intermediate values,
  especially for multi-step aggregation and comparison~\cite{wei2023chainofthoughtpromptingelicitsreasoning, akhtar-etal-2023-exploring}.
  \item \textbf{Faithfulness issues}: rationales may not correspond to actual table evidence (hallucinated), and answers
  can be sensitive to serialization order or irrelevant rows~\cite{yang2025causality}.
  \item \textbf{Context pressure}: Large or wide tables require truncation or retrieval to fit context windows~\cite{wang2024chainoftableevolvingtablesreasoning}. However, effective retrieval is inherently difficult over discrete table units (e.g., isolated cells or triples). 
  \item \textbf{Complex table structure}: multi-level headers, merged cells, and irregular layouts can be hard to linearize without
  losing semantics~\cite{cheng-etal-2022-hitab}.
\end{itemize}

\paragraph{2) Symbolic reasoning (Program-aided approaches).}
Symbolic reasoning explicitly produces an executable program (e.g., SQL, pandas/Python) whose execution yields the final answer:
\begin{equation}
\begin{split}
    \hat{p} &= \texttt{LLM}\big(q \,\Vert\, \texttt{Schema}(T)\big) \\
    \hat{a} &= \texttt{Exec}(\hat{p}, T)
\end{split}
\end{equation}
This paradigm includes classical semantic parsing (text-to-SQL) and modern LLM tool-use variants where the model generates
code and calls an executor iteratively~\cite{chen2023programthoughtspromptingdisentangling}.
Agentic variants extend this into multi-step tool interaction: the model alternates between reasoning, selecting an action
~\cite{zhang2023reactableenhancingreacttable,lu2025tartopensourcetoolaugmentedframework}. Similar autonomous agent paradigms, often driven by reinforcement learning, have also shown tremendous efficacy in navigating complex temporal structures in KGs~\cite{gong2026tempr1unifiedautonomousagent, gong2025rtqarecursivethinking}.
PoTable~\cite{mao2025potablesystematicthinkingstageoriented} further adopts a stage-oriented \emph{plan-then-execute} pipeline, where the LLM plans and generates Python operations
within analyst-inspired stages and iteratively corrects errors using real-time execution feedback.
Recent approaches also evolve intermediate tabular states in the reasoning chain (e.g., Chain-of-Table) so that each step
materializes partial results as a new table, reducing long-horizon reasoning burden~\cite{wang2024chainoftableevolvingtablesreasoning}.

\noindent\textbf{Strengths.}
Symbolic methods typically offer precision by delegating arithmetic and set operations to deterministic executors, while simultaneously ensuring verifiability through auditable program traces that allow for intermediate validation.

\noindent\textbf{Failure modes.}
However, symbolic reasoning can be brittle:
\begin{itemize}
  \item \textbf{Semantic inflexibility}: fixed operator libraries may not capture nuanced question intent, and LLM-generated programs
  may be logically incorrect or mishandle data; small mismatches in column naming or value formats/units can derail parsing
  and execution~\cite{cao2025tablemasterrecipeadvancetable,nahid-rafiei-2024-normtab}.
  \item \textbf{Sensitivity to complex tables}: irregular layouts, hierarchical headers, and implicit relationships can make schema
  extraction ambiguous; symbolic parsers may fail without careful canonicalization~\cite{cao2025tablemasterrecipeadvancetable}.
\end{itemize}

\paragraph{3) Hybrid reasoning (Textual $\oplus$ Symbolic).}
Hybrid systems integrate the semantic flexibility of textual reasoning with the precision of symbolic execution, typically employing paradigms such as \textbf{adaptive routing} for dynamic selection~\cite{liu2023rethinkingtabulardataunderstanding, zhang-etal-2024-syntqa}, and \textbf{interleaved modularity} for step-wise refinement~\cite{khoja-etal-2025-weaver} or failure-aware backtracking~\cite{liu2026cogcontrollablegraphreasoning} to mitigate hallucinations. Furthermore, bridging the intrinsic semantic gap between structured layouts and unstructured queries is a recognized prerequisite for complex reasoning, heavily explored in both TableQA and KGQA~\cite{Li_2025}. However, existing hybrid methodologies in TableQA are predominantly restricted to flat tables with canonical structures; they struggle to generalize to complex tables. In this work, we bridge this gap by introducing \textbf{AdaSTR}, which serializes complex tables into Semantic Trees. By normalizing irregular layouts into a structure with explicit hierarchy and unified semantics, AdaSTR successfully extends this high-precision hybrid paradigm (\textbf{DuTR}) to the domain of complex TableQA.

\section{Metrics and Mode Selection for Adaptive Tree Synthesis}
\label{appendix:mode_selection}
\subsection{Adaptive Mode Selection Policy}

Directly encoding large tables for LLM processing is often infeasible due to context-length constraints and the sharply increasing computational cost with table size (e.g., even a medium-sized table can exceed tens of thousands of tokens). We therefore adopt an automatic mode-selection policy that chooses between DSP/SRE/PSS according to a context budget and table statistics.

\paragraph{Token Budget and Lightweight Statistics.}
Instead of utilizing the entire physical context window $L_{\text{max}}$, we define an \textit{effective reasoning budget} $B$ to ensure robust attention performance. Let $L_{\text{safe}} = \alpha \cdot L_{\text{max}}$ be the safe context limit~\cite{liu2023lostmiddlelanguagemodels, hsieh2024rulerwhatsrealcontext}, and $L_{\text{sys}}$ be the system prompt cost. The available budget for table serialization is:
\begin{equation}
    B = L_{\text{safe}} - L_{\text{sys}} - \mathcal{E}
\end{equation}
where $\mathcal{E}$ estimates the reserved tokens for the synthesized tree structure. We estimate the table footprint using a lightweight serialization $\tau(T)$. Specifically, $\tau(T)$ transforms the table into a compact nested list format, which eliminates redundant separators to serve as a minimal-token baseline. We compute the following statistics: 
\begin{align}
    S &= \mathrm{Tok}(\tau(T)) \\
    \bar{s} &= \frac{1}{|C|}\sum_{c\in C}\mathrm{Tok}(c) \\
    r_{\text{long}} &= \frac{|\{c\in C : \mathrm{Tok}(c) > \gamma\}|}{|C|} \\
    n &= |\mathrm{rows}| \cdot |\mathrm{cols}|
\end{align}
where $S$ represents the estimated total token consumption of the table. To capture content density and scale, we define $\bar{s}$ as the average token length per cell, and $r_{\text{long}}$ as the long-cell ratio (the proportion of cells exceeding a length threshold $\gamma$). The variable $n$ serves as the table scale indicator, representing the total number of cells. Here, $C$ denotes the set of non-empty cells, and $\mathrm{Tok}(\cdot)$ is measured using the target model tokenizer.

\paragraph{Decision Rule.}
The selection logic prioritizes fitting the context first, then addressing verbose (text-dense) overflow, and finally handling massive-scale tables.
\begin{enumerate}
    \item \textbf{DSP (Default)}: If the estimated token footprint fits within the context budget, we use direct generation for maximum semantic fidelity.
    \begin{equation}
        S \le B 
    \end{equation}

    \item \textbf{SRE (Density-First)}: If the table exceeds the budget but the scale is not massive (cell count is manageable), we attribute the overflow primarily to verbose content (e.g., high long-cell ratio) and switch to Symbolic Reference Encoding for compression via address placeholders.
\begin{equation}
\begin{split}
    & S > B \ \wedge\ n \le n_{high} \\
    & \quad \wedge\ ( \bar{s}>\mu \ \vee\ r_{\text{long}}>\eta )
\end{split}
\end{equation}

    \item \textbf{PSS (Scale-First)}: If the table exceeds the budget and the number of cells is massive, we switch to Programmatic Structure Synthesis. PSS is most effective for hyperscale tables because loop-based code expands large structures more reliably than token-by-token enumeration.
    \begin{equation}
        S > B \ \wedge\ n > n_{high} 
    \end{equation}
\end{enumerate}
If $S > B$ but neither SRE nor PSS conditions are strictly met (a rare edge case), we default to SRE as the safer fallback to reduce context length while preserving structure.

\paragraph{Default Hyperparameters.}
Unless otherwise stated, we use $\alpha = 0.6$, $\mu=80$ tokens, $\eta=0.3$, and $n_{high}=1\times10^3$. We did not perform extensive hyperparameter tuning; thresholds were chosen once as reasonable defaults. In our ablation studies (Section~\ref{sec:ablation_studies}), enabling this policy improves evaluator scores (IC/SI) and, critically, reduces DSP-only failures caused by output truncation on oversized tables, thereby improving robustness.

\subsection{Evaluation Metrics for Semantic Tree Quality}
\begin{figure}[t] 
    \centering 
    \includegraphics[width=1\linewidth]{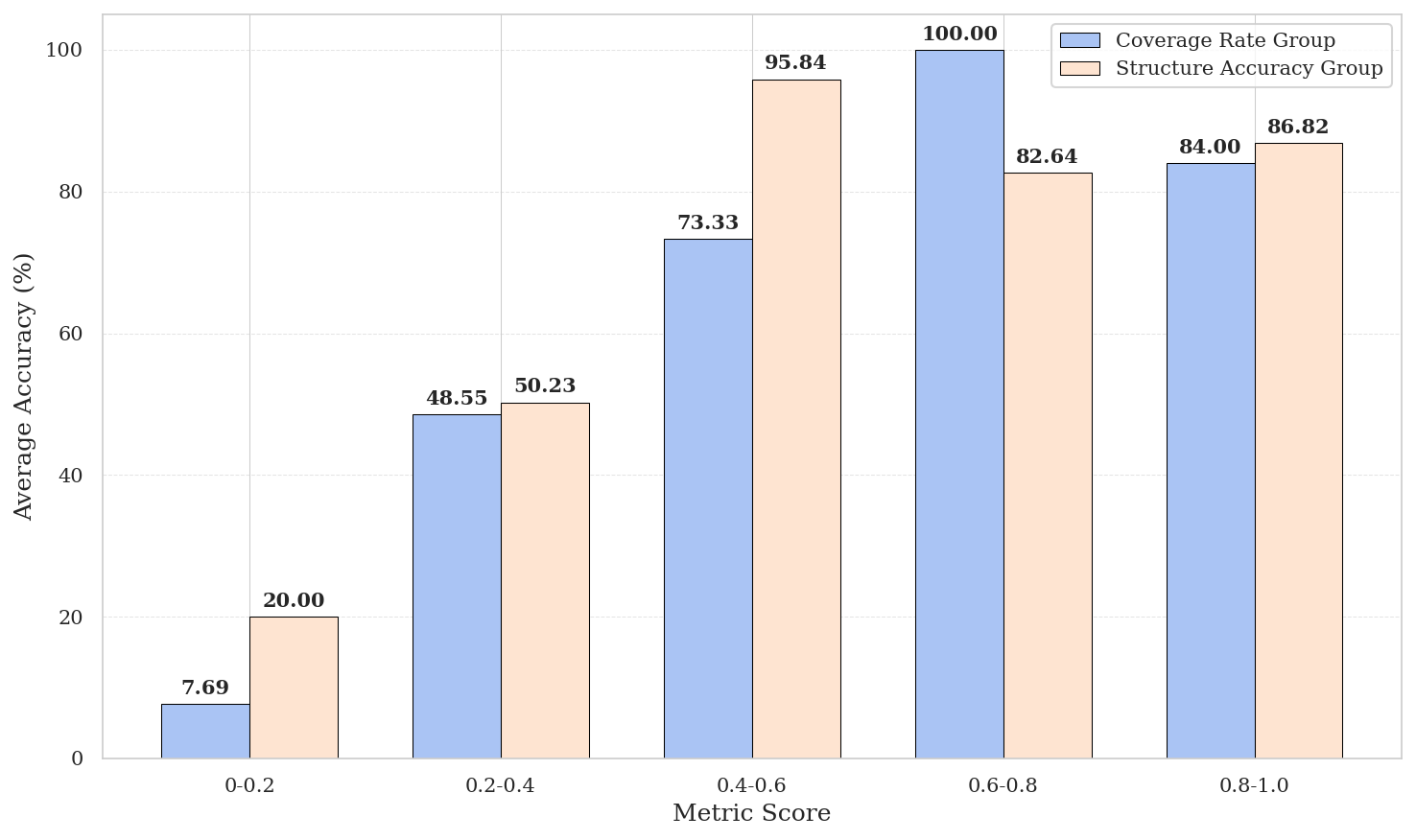} 
    \caption{Relationship between evaluation metric scores (information coverage and structure integrity) and downstream QA accuracy across score bins.
} 
    \label{fig:metric_analysis} 
\end{figure}

To rigorously assess the quality of the constructed semantic trees and guide the \textbf{Reflective Switching} mechanism, we define two quantitative metrics: \textbf{Information Coverage} and \textbf{Structural Integrity}. Although we do not provide a theoretical proof, our empirical observations (as illustrated in Figure~\ref{fig:metric_analysis}) demonstrate that these metrics serve as reliable proxies for downstream Question Answering (QA) performance.

\paragraph{1. Information Coverage.}
This metric measures the completeness of the information transfer from the tabular structure to the tree structure. It is calculated as the ratio of original table cells whose content is successfully represented in the generated tree nodes:
\begin{equation}
    \text{Coverage} = \frac{|C_{\text{mapped}}|}{|C_{\text{total}}|}
\end{equation}
where $C_{\text{mapped}}$ denotes the set of cells found in the tree and $C_{\text{total}}$ is the total number of non-empty cells in the original table.

\paragraph{2. Structural Integrity.}
This metric evaluates the correctness of the hierarchical relationships in the generated tree. We employ a bottom-up path verification strategy. For every data leaf node (value) in the tree, we trace the path back to the \texttt{ROOT}. The validity of a path is determined as follows:
\begin{itemize}
    \item \textbf{Initialization}: Start from the Leaf node. Let the Leaf be the distinct anchor.
    \item \textbf{Traversal}: Move upwards to the parent node.
    \item \textbf{Verification Logic}: Check the spatial alignment in the original table:
    \begin{enumerate}
        \item If the current node is in the \textbf{same row or same column} as the Leaf node, the relationship is valid; continue to the parent.
        \item If not, check if the current node is in the \textbf{same row or same column} as its immediate child node (the node just visited). If yes, the relationship is valid (transitive alignment); continue.
        \item If neither condition is met, the path is deemed \textbf{Structurally Broken}, and verification terminates.
    \end{enumerate}
    \item \textbf{Success}: If the traversal reaches the \texttt{ROOT} without error, the path is valid.
\end{itemize}
The \textit{Structural Integrity} score is the percentage of valid paths out of all leaf-to-root paths.

\paragraph{Discussion on Evaluation Metrics.} (1) \textbf{Merged Cell Representation:} Many table datasets lack explicit annotations for merged cell spans. For example, a header row [`Gender', `'] implies that the second cell inherits the label "Gender," yet the raw data often represents this as a null value. Consequently, during structural evaluation, items (e.g., "Male") nested under these implicit headers fail to align with their correct parent nodes, resulting in unwarranted penalties to the structural accuracy score.
(2) \textbf{Information Redundancy \& Normalization:} The AdaSTR method is designed to intelligently filter redundant information and normalize headers (e.g., factoring units out of cell values: 10 kg $\to$ Key: Weight (kg), Value: 10). While this restructuring offers superior semantic precision, it can trigger false negatives in our information coverage evaluation scripts, which may fail to recognize the data in its transformed state.

\paragraph{Correlation with Downstream Accuracy.}
To validate the proposed metrics, we analyzed their relationship with downstream task performance. As illustrated in Figure \ref{fig:metric_analysis}, we conducted correlation experiments on the SSTQA dataset. We sampled 200 instances and processed them exclusively using the SRE tree construction strategy. This specific strategy was selected to ensure that the metric scores were broadly distributed across the entire spectrum. For the downstream evaluation, we employed our proposed DuTR method to perform the question answering tasks. Our analysis reveals a significant positive correlation between the evaluation metrics and QA accuracy. This trend suggests that higher metric scores correspond to improved QA performance. Given that downstream accuracy serves as an intuitive proxy for tree quality, these results confirm the effectiveness of our evaluation framework. Notably, fluctuations in accuracy observed at higher metric scores are expected, as QA performance depends not only on tree structure but also on external factors such as question difficulty. Crucially, this empirical distribution guides the threshold configuration for our Evaluator-Guided Refinement Loop. We observe that when the \textbf{Coverage Rate} exceeds $0.8$ (the 0.8--1.0 bin), the average QA accuracy remains robust at $84.00\%$. Similarly, \textbf{Structure Accuracy} demonstrates a consistent performance plateau once it surpasses $0.6$, with the 0.8--1.0 bin achieving $86.82\%$. Based on these findings, we set the acceptance thresholds of quality assessment.

\paragraph{Metric-Guided Self-Correction.}
We implement a feedback loop that triggers specific refinement actions when the generated tree fails to meet pre-defined quality thresholds. The correction strategy is adaptive to the type of deficiency detected:
(1) \textbf{Structural Deficiency:} If the \textit{Structural Integrity} score is low, it indicates fundamental flaws in the logical hierarchy (e.g., incorrect parent-child relationships). In this case, we instruct the system to revert to the \textbf{Hierarchy Identification} phase to reconstruct the semantic backbone from scratch.
(2) \textbf{Information Omission:} Conversely, if the structure is valid but \textit{Information Coverage} is insufficient, it implies that the hierarchy is correct but specific data points are missing. Here, we retain the current tree and prompt the LLM to supplement missing information based on the raw table, while strictly maintaining the backbone hierarchy established in the Hierarchy Identification phase.

\section{Optimization Strategies for the Selector Module}
\label{app:selector_optimization}

The Selector module plays a critical role in determining the final output when the upstream generators produce conflicting answers. Thus, we conducted an in-depth analysis on a subset of 163 samples where the candidate answers were inconsistent. We investigate three distinct optimization strategies: Chain-of-Thought (CoT) reasoning, Supervised Fine-Tuning (SFT), and Logit-based Debiasing (Debias). To ensure deterministic reproducibility, we configured the Qwen3-8B selector with greedy decoding settings (temperature=$0$, do\_sample=False) and disabled its reasoning mode. Furthermore, for the SFT and Debiasing strategies, we directly utilized the model's output logits to determine the selected answer.

\paragraph{Baseline Performance.}
Our default implementation utilizes a direct Zero-shot Answer extraction prompt shown in Appendix~\ref{appendix:implementation_details}. On the conflicting subset, the baseline achieves an accuracy of 70.55\% (115/163), significantly outperforming random selection. The average inference latency is minimal ($\sim$0.5s), making it highly efficient.

\paragraph{Strategy I: Chain-of-Thought (CoT).}
We first attempted to enhance the reasoning capability of the Selector by incorporating Chain-of-Thought (CoT) prompts. The accuracy improved marginally to 72.39\% (118/163), a gain of only 3 correct samples. However, the inclusion of reasoning steps caused a drastic spike in latency, increasing from 0.5s to $\sim$13s on average. This suggests that for the specific task of binary/multiple choice selection, the computational overhead of CoT yields diminishing returns. The small-scale model may struggle to effectively leverage the generated reasoning path to correct subtle errors.

\paragraph{Strategy II: Supervised Fine-Tuning (SFT).}
To better align the model with the selection task, we constructed a specialized SFT dataset derived from the HiTab training set. 
\textbf{Data Construction:} We generated 1,000 training samples by pairing the ground truth with a "distractor" answer (a plausible but incorrect answer generated by the LLM). To prevent the model from memorizing answer positions, we applied data augmentation by swapping the order of correct and incorrect answers with a probability of 0.7, resulting in a total of 2,700 samples. 
\textbf{Training Setup:} We fine-tuned the Qwen3-8B model using QLoRA (4-bit quantization) to ensure efficiency~\cite{NEURIPS2023_1feb8787}. Key hyperparameters included a LoRA rank of $r=16$ targeting all linear layers, a learning rate of $2.0 \times 10^{-5}$ with a cosine scheduler (0.1 warmup ratio), and an effective batch size of 8 trained for 3 epochs using BF16 precision.
\textbf{Result \& Analysis:} After fine-tuning, the accuracy rose significantly to \textbf{76.68\%} (125/163). The substantial improvement indicates that the Selector benefits more from learning the specific error patterns and task format through parameter updates than from zero-shot reasoning.

\paragraph{Strategy III: Logit-based Debiasing (Debias).}
Recent studies~\cite{zheng2023judgingllmasajudgemtbenchchatbot} show that LLMs exhibit Positional Bias, often preferring the first option (Option A) in multiple-choice tasks. We implemented a parallel decoding strategy to mitigate this without training. 
\textbf{Method:} We feed the input twice in parallel with the candidate answers swapped (Order $A,B$ and Order $B,A$). Instead of relying on generated text, we sum the distinct token logits for each candidate across both permutations to determine the winner. 
\textbf{Result:} This method achieved an accuracy of 73.61\% (120/163). While slightly less effective than SFT, this approach requires no training and maintains low latency ($\sim$0.6s via parallel batching), serving as a cost-effective alternative.

\paragraph{Summary.}
Table~\ref{tab:selector_ablation} summarizes the trade-offs. We conclude that SFT offers the highest performance ceiling, while the Baseline or Logit-based methods provide the best efficiency.

\begin{table}[h]
\centering
\small
\resizebox{\linewidth}{!}{
\begin{tabular}{lccc}
\toprule
\textbf{Method} & \textbf{Accuracy} & \textbf{Latency} & \textbf{Training} \\
\midrule
Baseline (Direct) & 70.55\% & $\sim$0.5s & No \\
Strategy I (CoT) & 72.39\% & $\sim$13.0s & No \\
Strategy II (SFT) & \textbf{76.68\%} & $\sim$0.5s & Yes \\
Strategy III (Debias) & 73.61\% & $\sim$0.6s & No \\
\bottomrule
\end{tabular}
}
\caption{Performance comparison of optimization strategies on the conflicting subset ($N=163$).}
\label{tab:selector_ablation}
\end{table}

\section{Error Analysis}

To understand the limitations of our Dual-Mode Tree Reasoning framework, we conducted a failure analysis on a stratified random sample of $N=60$ error cases drawn from the SSTQA dataset. We sampled equally from three distinct failure populations ($n=20$ each): (1) cases where both Symbolic and Textual modes failed, (2) cases where Textual succeeded but Symbolic failed, and (3) cases where Symbolic succeeded but Textual failed. 

\begin{table*}[t]
\centering
\small
\begin{tabular}{l p{7cm} c c}
\toprule
\textbf{Error Type} & \textbf{Definition} & \textbf{Count} & \textbf{\%} \\
\midrule
\textbf{Structural Misalignment} & The generated code assumes a rigid schema (e.g., specific depth, key-value placement) that does not match the heterogeneous tree structure, leading to execution failures. & 22 & 36.7\% \\
\textbf{Annotation Error} & The model output is factually correct given the table content, but the gold label is incorrect due to annotation errors, typos, or calculation mistakes in the dataset. & 16 & 26.7\% \\
\textbf{Arithmetic Hallucination} & Textual reasoning fails at precise operations: arithmetic aggregation, numerical comparison (especially negative numbers/decimals). & 12 & 20.0\% \\
\textbf{Retrieval Bias} & The model exhibits \textit{semantic short-circuiting}, selecting high-level summary nodes (e.g., "Total") that match keywords but lack the required categorical granularity, failing to drill down to specific leaf details. & 10 & 16.7\% \\
\bottomrule
\end{tabular}
\caption{Distribution of error types across $N=60$ sampled failure cases.}
\label{tab:error_dist}
\end{table*}

\subsection{Error Taxonomy and Distribution}

We categorized the errors into four primary types based on the root cause analysis (Table~\ref{tab:error_dist}).

\begin{figure*}[t] 
    \centering 
    \includegraphics[width=1\linewidth]{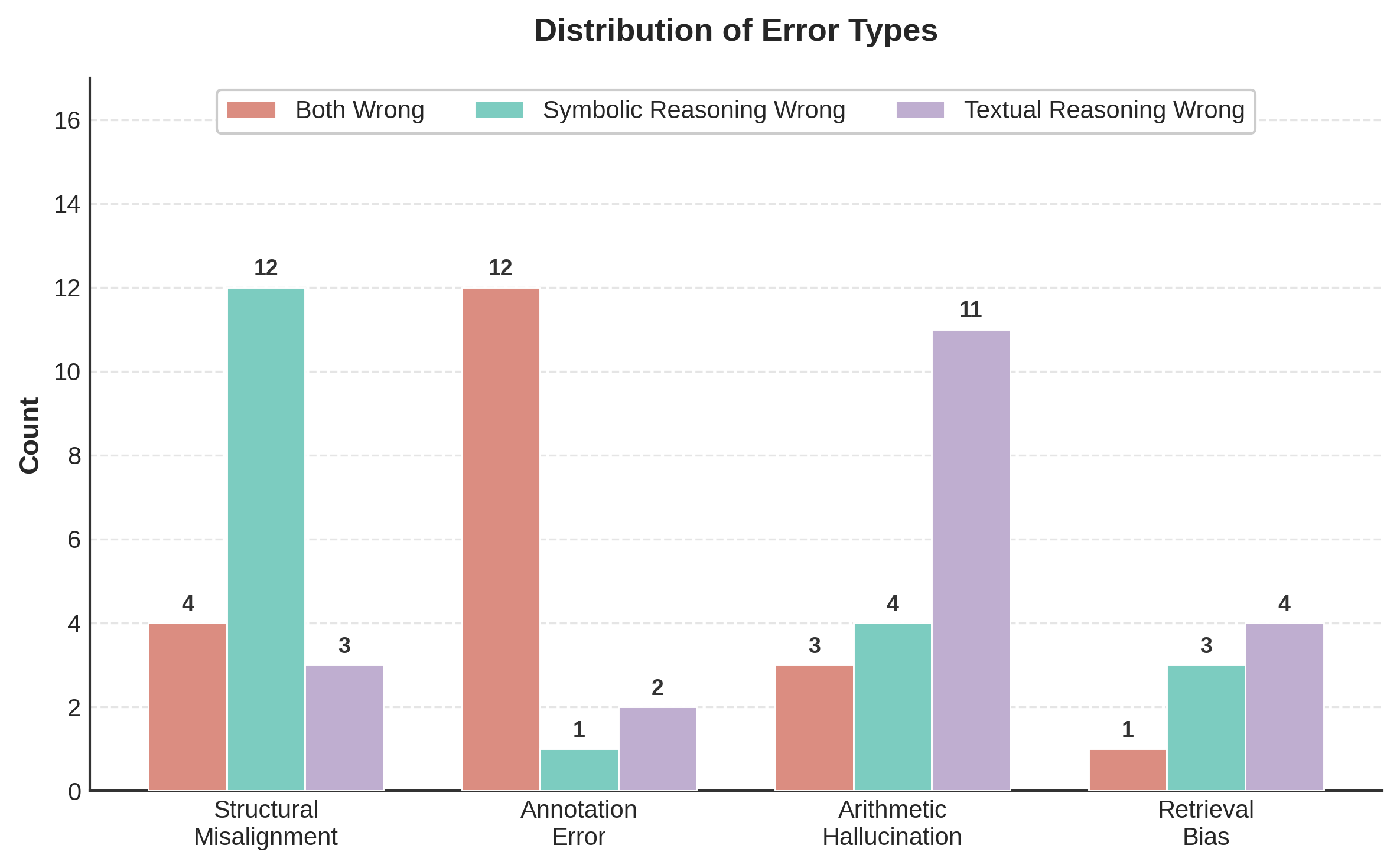} 
    \caption{Statistics of error categories by reasoning failure mode.
} 
    \label{fig:error_distribution} 
\end{figure*}

As illustrated in Figure~\ref{fig:error_distribution}, the distribution of error types varies significantly across the failure populations:
\begin{itemize}
    \item \textbf{Mode-Specific Weaknesses:} There is a clear decoupling of errors. \textit{Structural Misalignment} is the dominant failure mode for Symbolic Reasoning (12 cases), confirming that code generators struggle with rigid schema assumptions. Conversely, \textit{Arithmetic Hallucination} is heavily skewed towards Textual Reasoning (11 cases), highlighting the LLM's inability to perform precise calculations without external tools.
    
    \item \textbf{The Consensus Insight:} The \textit{Annotation Error} category shows a unique pattern: it is overwhelmingly composed of "Both Wrong" cases (12 out of 15). This indicates that when our Textual and Symbolic modules reach a consensus that disagrees with the gold label, the discrepancy is usually due to dataset flaws (e.g., wrong ground truth) rather than model error.
\end{itemize}
\subsection{Analysis of Representative Failures}

\paragraph{1. Structural Misalignment (36.7\%).}
While this error affects both paradigms, it is predominantly a Symbolic failure (12 cases vs. 3 Textual).
The Symbolic module struggles because code generators often hallucinate a flattened schema for deeply nested trees (e.g., missing a parent key like \texttt{Loan Bank} and iterating directly over children). 
\textit{Textual reasoning is more robust} to structure because it reads the serialized tree linearly, though it occasionally fails (3 cases) when long context windows obscure the indentation levels of remote leaf nodes. This contrast highlights that while LLMs understand serialized semantics (Textual) well, mapping them to rigid executable paths (Symbolic) remains fragile without agentic inspection.

\paragraph{2. Annotation Errors (26.7\%).}
This category represents a mode-agnostic failure, heavily concentrated in the "Both Wrong" intersection (12 cases).
Unexpectedly, a quarter of failures were actually correct model predictions penalized by flawed datasets. For instance, given the query \textit{"variance rate"}, both models correctly extracted the percentage (-1.4), but the gold label (-2800) referred to the absolute difference.
The fact that \textit{both} reasoning modes—despite their orthogonal mechanisms—arrived at the same answer contradicting the label, suggests that system consensus is often a stronger signal of truth than the noisy ground truth labels.

\paragraph{3. Arithmetic \& Logic Hallucination (20.0\%).}
This error manifests in distinct forms across the two modes.
For \textbf{Textual Mode} (11 cases), it is a \textit{Calculation Failure}: the model retrieves correct numbers but drifts during long-chain floating-point aggregation (e.g., $26.8 + \dots$) due to the lack of a calculator.
However, \textbf{Symbolic Mode} is not immune (4 cases); it suffers from \textit{Formula Logic Hallucination}. Although the Python execution is flawless, the model may generate the wrong operator (e.g., \texttt{sum} instead of \texttt{mean}) or fail to normalize units (e.g., treating "1,000" as a string), leading to numerically wrong answers. Thus, Symbolic reasoning solves calculation but introduces logic derivation risks.

\paragraph{4. Retrieval Bias (16.7\%).}
This error reflects a "granularity mismatch" where models select high-level summary nodes instead of specific leaves.
It is more prevalent in Textual Mode (4 cases) due to ``semantic short-circuiting''—the model stops reading at the first keyword match (e.g., "Total Expenditure") ignoring temporal constraints.
\textbf{Symbolic Mode} also encounters this (3 cases), but for a different reason: \textit{Naming Ambiguity}. If a parent node and a child node share similar key names, the code generator may ambiguously reference the parent object instead of drilling down, failing to satisfy the specific granularity required by the question.

\section{Additional Experimental Analysis}
\label{sec:additional_experiments}

\subsection{Comparison with Relational Transformation Methods}
\label{appendix:relational_transformer}

As illustrated in Table~\ref{tab:Realtional_table_comp}, to ensure a fair comparison with \textbf{RelationalCoder} and \textbf{TabFormer}, both of which utilize \textbf{GPT-3.5} as the reasoning backbone and report results using \textit{Exact Match (EM)} on the HiTab dataset, we align our experimental setup with these established protocols by employing \textbf{GPT-3.5} as the downstream \textit{QA reasoner}. Crucially, our primary objective is to investigate which intermediate representation format—our Logical Semantic Tree versus baseline relational or linear formats—more effectively enhances the reasoning capabilities of LLMs. By standardizing the downstream reasoning engine to GPT-3.5, we isolate the data representation as the key variable. Thus, while we utilize DeepSeek-V3 for the upstream serialization (AdaSTR), the performance difference observed under the same reasoner directly highlights the superior information accessibility and structural clarity provided by our tree representation.

\begin{table}[h]
\centering
\resizebox{\linewidth}{!}{
\begin{tabular}{llc}
\toprule
\textbf{Structure Representation} & \textbf{Reasoning Method} & \textbf{EM (\%)} \\
\midrule
\multicolumn{3}{l}{\textit{\textbf{Baselines}}} \\
\addlinespace[3pt]
\multirow{3}{*}{TabFormer\cite{10.1145/3726302.3730071}} & Chain-of-Table\cite{wang2024chainoftableevolvingtablesreasoning} & 57.4 \\
                           & E5\cite{zhang-etal-2024-e5}   & 56.9 \\
                           & ReactTable\cite{zhang2023reactableenhancingreacttable}     & 54.3 \\
\cdashline{1-3} 
\addlinespace[2pt] 
RelationalCoder\cite{dong-etal-2025-relationalcoder}            & Chain-of-Table\cite{wang2024chainoftableevolvingtablesreasoning} & 66.7 \\

\midrule
\addlinespace[3pt]
\multirow{4}{*}{\textbf{AdaSTR (Ours)}} & DuTR(textual reasoning) & 64.1 \\
& \textbf{DuTR(symbolic reasoning)} & \textbf{74.1} \\
& DuTR(adaptive selection) & \underline{68.9} \\
& \textit{DuTR(Oracle)} & 80.4 \\
\bottomrule
\end{tabular}
}
\caption{Supplementary comparison with relational transformation methods. We evaluate different structure representations coupled with various reasoning methods.}
\label{tab:Realtional_table_comp}
\end{table}

Experimental results demonstrate that our \textbf{DuTR (Symbolic reasoning)} achieves an EM score of $74.1\%$ on HiTab, significantly outperforming the strongest relational baseline, RelationalCoder ($66.7\%$). This performance disparity highlights a fundamental divergence in data representation. RelationalCoder and TabFormer attempt to flatten or decompose nested complex tables into standardized relational tables---a transformation aligned with SQL logic. However, this process disrupts the original \textit{Physical Locality} and \textit{Hierarchical Context}. Consequently, the model incurs a higher cognitive load during reasoning, as it must implicitly execute complex ``Join'' operations to reconstruct cross-level relationships.

In contrast, our \textbf{Semantic Tree} preserves the topological integrity of the data. Parent and child nodes are directly connected, allowing the LLM to perform aggregation via recursive operations on local substructures without the need for cross-table retrieval. This \textit{Structure-Logic Isomorphism} is pivotal to the observed performance gains. On the computation-intensive HiTab dataset, our Symbolic Tree Manipulation yields optimal performance. Notably, our method maintains a substantial lead even when deployed on the relatively weaker GPT-3.5 backbone. This suggests that the performance gains are derived primarily from superior data structure design rather than solely from the intrinsic capabilities of a stronger model.

\textbf{Analysis of Adaptive Mode Performance:} The slightly lower performance of the Adaptive mode compared to the Symbolic mode can be attributed to the rigidity of the Exact Match (EM) metric. EM demands character-level correspondence between the output and the reference answer. Since we did not perform granular optimization to align the textual output with specific EM formatting constraints, the Adaptive selector occasionally chooses responses that are semantically correct but syntactically mismatched (i.e., failing the EM check). This phenomenon introduces false negatives, thereby penalizing the overall score of the Adaptive strategy.

\subsection{Generalization on Open-Source Models}
\label{subsec:sstqa_opensource}
As shown in Table~\ref{tab:sstqa_opensource_transposed_grouped}, the Semantic Tree representation generalizes well across different open-source backbones and inference settings. Compared with directly prompting models with the original raw tables, converting the input into a Tree Table consistently improves performance for every model tested (e.g., +3.14 for Qwen2.5-7B, +1.44 for Llama3.1-8B, and up to +9.95 for Qwen3-8B in thinking mode). This suggests that providing an explicitly structured, hierarchy-aware layout reduces surface-form ambiguity and better aligns with models' internal reasoning patterns. We further observe that \textbf{symbolic-only} inputs can be brittle for some open-source models, most notably Llama3.1-8B (10.47), indicating that purely symbolic signals without sufficient linguistic grounding may be difficult to interpret reliably.

\begin{table}[t]
\centering
\resizebox{\columnwidth}{!}{%
\begin{tabular}{lcccc}
\toprule
\multirow{2}{*}{Method} & \multirow{2}{*}{Qwen2.5-7B} & \multirow{2}{*}{Llama3.1-8B} & \multicolumn{2}{c}{Qwen3-8B} \\
\cmidrule(lr){4-5}
 &  &  & no thinking & thinking \\
\midrule
Direct Prompt (Raw Table) & 46.60 & 35.99 & 52.75 & 54.97 \\
Direct Prompt (Semantic Tree)      & 49.74 & 37.43 & \underline{60.86} & 64.92 \\
\textbf{ASTRA} (Textual Reasoning)   & \underline{50.14} & \underline{40.45} & 60.60 & \underline{65.45} \\
\textbf{ASTRA} (Symbolic Reasoning)  & 36.39 & 10.47 & 58.64 & 62.70 \\
\textbf{ASTRA} (Adaptive Selection)  & \textbf{53.01} & \textbf{41.49} & \textbf{65.31} & \textbf{70.02} \\
\textbf{ASTRA} (Oracle)              & 60.64 & 44.76 & 75.26 & 78.14 \\
\bottomrule
\end{tabular}%
}
\caption{Performance of Open-Source Models on the SSTQA Dataset. The best results are highlighted in \textbf{bold}, and the second-best results are \underline{underlined}. Note that the ``Direct Prompt'' method follows the same workflow as the Foundation model approach used in the main experiments.}
\label{tab:sstqa_opensource_transposed_grouped}
\end{table}

\begin{table}[t]
\centering
\resizebox{\linewidth}{!}{%
\begin{tabular}{lcc}
\toprule
\textbf{Configuration} & \textbf{Acc (\%)} & \textbf{$\Delta$} \\
\midrule
\textbf{Adaptive Tree Navigation} & \textbf{60.60} & - \\
\quad w/o Embedding Model & 56.68 & \textcolor{red}{-3.92} \\
\quad w/o Dynamic Switching & - & - \\
\quad \quad \textit{Force Root-to-Leaf} & 55.76 & \textcolor{red}{-4.84} \\
\quad \quad \textit{Force Leaf-to-Root} & 58.64 & \textcolor{red}{-1.96} \\
\midrule
\textbf{Symbolic Tree Manipulation} & \textbf{58.64} & - \\
\quad w/o Code Examples & 50.89 & \textcolor{red}{-7.75} \\
\quad w/o Structural Skeleton & 53.12 & \textcolor{red}{-5.52} \\
\quad w/o Self-Correction Loop & 45.97 & \textcolor{red}{-12.67} \\
\midrule
\multicolumn{3}{l}{\textit{\textbf{Impact of Data Representation}}} \\
\quad \textit{Raw Table (Direct Prompting)} & 52.75 & - \\
\quad \textit{Semantic Tree (Direct Prompting)} & \textbf{60.86} & \textcolor{good}{+8.11} \\
\bottomrule
\end{tabular}%
}
\caption{Ablation analysis of \textbf{DuTR (Qwen3-8B)}. We report the Accuracy (\%) and the performance change ($\Delta$) compared to the full configuration within each mode.}
\label{tab:ablation_treeqa}
\vspace{-2mm}
\end{table}

The comparison between Qwen3-8B with and without thinking shows that stronger inference-time reasoning amplifies the benefit of structured representations: thinking mode improves Tree Table prompting from 60.86 to 64.92 (+4.06), reinforcing that our Semantic Tree guidance remains effective even when the model's reasoning strategy changes. 

To further investigate module contributions on smaller parameters, we conducted a detailed ablation study on Qwen3-8B (Table~\ref{tab:ablation_treeqa}). A striking contrast with the larger DeepSeek-V3 model emerges in symbolic reasoning: while DeepSeek-V3 showed resilience to removing the \textbf{Self-Correction Loop} (dropping only $1.57\%$), Qwen3-8B suffered a severe degradation ($-12.67\%$). This indicates that while frontier models possess robust intrinsic code generation capabilities, smaller models rely critically on iterative execution feedback to rectify syntax or logic errors. Additionally, the \textbf{Static Tree} representation alone yielded a \textbf{+8.11\%} gain over raw tables, confirming that the cognitive scaffold provided by our tree structure remains beneficial independent of model scale.

Overall, these results demonstrate that the proposed representation is not tailored to any single proprietary model, but instead serves as a robust interface for open-source LLMs with varying parameter scales and reasoning capabilities.

\section{Case Study}
\label{app:case_study}

In Section~\ref{sec:case_study}, we demonstrated how ASTRA addresses \textit{Structural Neglect} and \textit{Representation Gap} through a representative case on hierarchical enumeration (\textbf{Explicit Hierarchy}). 
To provide a comprehensive evaluation, we present two additional case studies in this appendix, focusing on \textbf{Representation Alignment} (handling null/missing values) and \textbf{Symbolic Compatibility} (performing long-horizon arithmetic). These cases, summarized in Table~\ref{tab:case_study_null_aggregation} and Table~\ref{tab:case_study_arithmetic}, further validate the necessity of our Text-Symbolic paradigm.

\begin{table*}[t]
    \centering
    \small
    \renewcommand{\arraystretch}{1.4} 
    \begin{tabular}{p{0.15\linewidth} | p{0.8\linewidth}}
        \toprule
        \rowcolor{bgGray}
        \textbf{Question} & \textbf{How many expense records have an empty summary?} \\
        \rowcolor{bgGray}
        \textbf{Gold Answer} & \textbf{\textcolor{cGreen}{26}} \\
        \midrule
        
        \textbf{GraphOTTER} & 
        \textbf{\textcolor{cRed}{Output: 4}} \newline
        \textbf{Evidence:} The dense retriever only located 4 isolated empty cells out of 26: \newline
        \texttt{[(5, 6, `'), (13, 6, `'), (16, 6, `'), (19, 6, `')]} \newline
        \textit{Failure Analysis:} The vector-based retriever struggled to match the query "empty summary" with actual empty strings, resulting in extremely low recall. \\
        \midrule
        
        \textbf{ST-Raptor} & 
        \textbf{\textcolor{cRed}{Output: 46}} \newline
        \textbf{Process:} Decomposed the query and generated the operation chain: \texttt{[MATH] + [Summary] + [CNT]}. \newline
        \textit{Failure Analysis:} While it correctly identified the column and the counting task, the execution module failed to apply the "is empty" filter, effectively counting \textit{all} 46 rows in the table instead of the specific subset. \\
        \midrule
        
        \textbf{EEDP} & 
        \textbf{\textcolor{cRed}{Output: 29}} \newline
        \textbf{Process:} Attempted a manual row-by-row inspection in the CoT: \newline
        \textit{"The following rows have null... 004, 012, 015, 017..."} \newline
        \textit{Failure Analysis:} The model hallucinated the existence of empty values in non-empty rows during the text generation process, leading to an arithmetic error (counting 29 instead of 26). \\
        \midrule
        
        \textbf{Ours} & 
        \textbf{\textcolor{cGreen}{Output: 26}} \newline
        \textbf{Logic:} synthesized a programmatic constraint to filter data accurately: \newline
        \texttt{\textcolor{cBlue}{if not} record[\textquotesingle Summary\textquotesingle].strip(): count += 1} \newline
        \textit{Result:} The symbolic execution over the structured tree yielded the exact count (26) without hallucination. \\
        \bottomrule
    \end{tabular}
    \caption{\textbf{Case Study on Null Value Aggregation.} Comparison of failure modes: GraphOTTER suffers from \textit{retrieval miss} on empty values (poor \textit{Representation Alignment}); ST-Raptor correctly plans but fails in \textit{execution} (returning total count); EEDP suffers from \textit{hallucination} during manual enumeration. Our method ensures correctness via symbolic code generation.}
    \label{tab:case_study_null_aggregation}
\end{table*}

\subsection{Addressing Linguistic Misalignment in Retrieval}
\label{subsec:case_linguistic}
Queries involving negative constraints or null values (e.g., \textit{"empty summary"} in Table~\ref{tab:case_study_null_aggregation}) expose the failure of current serializations to achieve Representation Alignment. Existing methods like GraphOTTER convert tables into triples or vectors that fail to preserve the natural language context of "missingness," causing the retriever to miss empty cells entirely (Representation Gap). Similarly, EEDP's linear serialization lacks distinct markers for null values, leading to hallucination during row scanning. In contrast, our framework achieves Symbolic Compatibility by synthesizing programmatic constraints (\texttt{if not record[`Summary']}). By shifting from ambiguous linguistic matching to precise symbolic logic, we accurately identify all 26 records, validating the necessity of supporting verifiable code-based reasoning over purely textual serialization.
\subsection{Eliminating Arithmetic Hallucination (Re: Symbolic Compatibility)}
\label{subsec:case_arithmetic}
Long-horizon numerical reasoning (Table~\ref{tab:case_study_arithmetic}) critically tests Symbolic Compatibility. While baselines may retrieve the correct data, they fail to satisfy the requirement for verifiable computation. As observed with GraphOTTER and EEDP, feeding serialized floating-point numbers directly into an LLM triggers arithmetic hallucination ($48,154 \neq 51,596$), as the text-based modality is ill-suited for high-precision calculation. Our approach fulfills the symbolic requirement by offloading computation to a Python environment. The generated code naturally handles data type filtering and performs operations with machine precision, demonstrating that an ideal serialization must support the seamless transition from textual representation to executable logic.

\begin{table*}[t]
    \centering
    \small
    \renewcommand{\arraystretch}{1.4}
    \begin{tabular}{p{0.15\linewidth} | p{0.8\linewidth}}
        \toprule
        \rowcolor{bgGray}
        \textbf{Question} & \textbf{What is the average standard error of product sales from month 1 to month 18?} \\
        \rowcolor{bgGray}
        \textbf{Gold Answer} & \textbf{\textcolor{cGreen}{3685.4505}} \\
        \midrule
        
        \textbf{GraphOTTER} & 
        \textbf{\textcolor{cRed}{Output: 3439.61}} \newline
        \textbf{Evidence:} The model correctly identified all 14 relevant cells (Months 5--18) and recognized the operation as "Average". \newline
        \textit{Failure Analysis:} \textbf{Arithmetic Hallucination}. Despite correct retrieval, the LLM failed to sum the sequence of 14 floating-point numbers accurately during text generation, resulting in an incorrect sum ($48,154$ instead of $\approx51,596$) and final average. \\
        \midrule
        
        \textbf{ST-Raptor} & 
        \textbf{\textcolor{cRed}{Output: 4287.34}} \newline
        \textbf{Process:} Decomposed into \texttt{[COND] (Month 1-18) -> [MATH] + [AVR]}. \newline
        \textit{Failure Analysis:} \textbf{Execution Error}. While the retrieval and plan were semantically correct, the mathematical primitive likely mishandled the `None` values present in Months 1--4 or failed to compute the weighted mean of the float values correctly. \\
        \midrule
        
        \textbf{EEDP} & 
        \textbf{\textcolor{cRed}{Output: 3399.74}} \newline
        \textbf{Process:} The CoT explicitly listed the 14 values to sum: \textit{"3355.12 + 3220.81 + ..."} \newline
        \textit{Failure Analysis:} \textbf{Calculation Error}. Similar to GraphOTTER, the Chain-of-Thought process suffered from calculation drift. The model calculated the sum as $47,596.35$, significantly deviating from the true sum, leading to an underestimated average. \\
        \midrule
        
        \textbf{Ours} & 
        \textbf{\textcolor{cGreen}{Output: 3685.4505}} \newline
        \textbf{Logic:} Generated a symbolic program to iterate and aggregate valid data: \newline
        \texttt{\textcolor{cBlue}{if} error \textcolor{cBlue}{is not None}: total += error; count += 1} \newline
        \textit{Result:} By offloading the computation to a Python execution environment, our method achieved exact numerical precision, correctly handling null filtering and floating-point arithmetic. \\
        \bottomrule
    \end{tabular}
    \caption{\textbf{Case Study on Arithmetic Aggregation.} Comparison of failure modes on long-sequence numerical queries. Baselines correctly retrieve data but suffer from \textit{arithmetic hallucination} or \textit{primitive execution errors}, proving the inadequacy of textual processing. Our method ensures verifiable precision via code execution, fulfilling \textbf{Symbolic Compatibility}.}
    \label{tab:case_study_arithmetic}
\end{table*}



\clearpage
\newpage
\onecolumn
\section{Prompt}

\begin{lstlisting}[caption={AdaSTR: Header Normalization}]
You are a data analysis expert specializing in parsing complex table structures.

Your task is to analyze the first few rows of the provided table and determine a unique normalized header for each column. You must strictly follow these rules:

1. **Strict Preservation of Original Wording**: You must preserve the original text from the table cells. It is forbidden to create new names or summaries.
   (For example: if a cell contains "Total Students", the header must include "Total Students", not "Total Students Metrics".)
2. **Hierarchical Combination**: If header information is distributed across multiple header rows, combine them using the format: [Lower-Level Header] - [Upper-Level Header].
   The part before " - " must be the most specific level.
3. **Strict Column Count Matching**: The number of strings in the output array must exactly match the number of columns in the data rows.

The table is provided in JSON array form.

[Input Table]:
{TABLE_AS_JSON_STRING}

Your output must be a single, valid JSON string array representing the normalized headers. Do not provide any explanation.
\end{lstlisting}

\begin{lstlisting}[caption={AdaSTR: Hierarchy Identification}]
You are a data architecture expert specializing in analyzing the logical structure of tables.

You will be given a table and its normalized headers. Your task is to identify which columns belong to "Hierarchy Keys" and which belong to "Value Leaves".

You must reason about the table step by step and then produce a single JSON object as the final answer.

## Step 1: Inspect column roles

For each column, think about:
- Is it mostly numeric (measures), mostly text (categories), or mixed?
- Roughly how many distinct values does it have compared to the number of rows?
- Do the values look like IDs / timestamps / codes (often unique), or like categories (heavily reused)?

Use these heuristics:
- Columns that are mostly numeric or percentages are usually **value_leaves**.
- Columns whose values repeat a lot (far fewer unique values than rows) are strong candidates for **hierarchy_keys**.
- A column whose values are almost all unique (ID-like or timestamp-like) is a good **primary hierarchy_key** for simple flat tables.

## Step 2: Detect header-based hierarchies (multi-row headers)

If the original table had multiple header rows (already compressed into the normalized headers you see), then:
- When a normalized header clearly encodes multiple levels (e.g. "Sales - 2023 - Q1"), its column is usually a **value_leaf** grouped under a higher-level semantic group.
- Semantic group names must come directly from the original headers; do not invent new group names.

## Step 3: Detect row-based block structures (very important for complex tables)

Even if there is only one obvious text column, the table may still be **complex** if it uses row blocks.

Treat the table as **complex** (not simple) when you observe patterns like:

1. **Block headers + detail rows in the same column**:
   - Some rows in a text column look like high-level categories and have many empty cells or missing numeric values in other columns.
   - The rows immediately following such a row contain repeated labels such as "total", "weekday", "weekend", etc. with numeric values filled.
   - The same small set of labels ("total", "weekday", "weekend", ...) appears multiple times in separated blocks.

2. **Strong repetition of small categorical sets**:
   - A column alternates between "group header" values and a small fixed set of "detail" values in a patterned way.
   - This usually indicates a hierarchy like: Behaviour Group > Behaviour Detail.

In such cases:
- You should set "table_type" to "complex", even if there is only one text column.
- The **hierarchy_keys** should include that text column.

## Step 4: Decide simple vs complex

Use these refined rules:

- Classify as **simple** only if ALL of the following hold:
  1. There is no obvious multi-row header structure.
  2. There is no clear row-based block pattern as described in Step 3.
  3. Each row is largely independent.
  4. Typically there is at most one main categorical column and the others are mostly numeric measures.

- Classify as **complex** if ANY of the following holds:
  1. There are multiple categorical columns that naturally form a chain.
  2. Some categorical columns have heavily repeated values and clearly act as grouping keys.
  3. There are apparent aggregation / subtotal / summary rows.
  4. There is a row-based block structure as described in Step 3.

## Output Requirements

Your output must be a single, valid JSON object, and must strictly follow the structure below:

{
  "table_type": "simple" or "complex",
  "analysis_reason": "A brief explanation of the reasoning behind your judgment",
  "hierarchy_keys": ["header1", "header2", ...],
  "value_leaves": ["header3", "header4", ...],
  "semantic_groups": {
    "OriginalGroupName1": ["header_a", "header_b"]
  }
}

Additional constraints:
- Every header in "hierarchy_keys" and "value_leaves" must come from the normalized header list.
- The union of "hierarchy_keys" and "value_leaves" must exactly cover all normalized headers.
- If "table_type" is "simple", "hierarchy_keys" should usually contain only **one** element.

[Input Table]:
{TABLE_AS_JSON_STRING}

[Normalized Headers]:
{NORMALIZED_HEADERS_FROM_STEP_1}
\end{lstlisting}

\begin{lstlisting}[caption={AdaSTR: Tree Construction}]
You are a highly specialized data transformation engine responsible for converting tabular data into a nested JSON tree that preserves all semantic information.

You will be given the original table, its normalized headers, and the hierarchy definition. You must construct the final JSON tree based on these inputs, strictly following the rules below.

Before constructing the tree, you MUST:
- Read the "table_type" and "analysis_reason" from the hierarchy definition.
- Decide whether to use a flat structure (for simple tables) or a nested structure (for complex tables).
- When the analysis_reason mentions row-based or block hierarchy, you must reconstruct that hierarchy using the row patterns as well as the hierarchy_keys.

1. Semantic Hierarchy Keys
- For each hierarchy_key column, the corresponding key in the JSON must follow the format "[Header Name] - [Cell Value]".
- You must use " - " (space, hyphen, space) as the separator.
- Example: if the header is "Grade" and the cell value is "1", then the JSON key should be "Grade - 1".

2. Strict Preservation of Leaf Node Names
- The keys of value_leaves in the final JSON must exactly match the normalized headers.

3. General Structure Generation Rules
- Traverse the data row by row, skipping the header row.
- For each data row, use the hierarchy_keys (in order) to create or move into the corresponding nested structure.
- At the deepest level for a given row, create an object that stores all value_leaves for that row.

4. Handling of Simple Tables (table_type = "simple")
For simple tables, you should keep the structure shallow and flat:
- Typically there is only one hierarchy_key.
- For each data row:
  - Construct a single key by applying the "[Header Name] - [Cell Value]" rule to that hierarchy_key column.
  - Store an object under that key with all value_leaves for that row.

5. Handling of Complex Tables (table_type = "complex")

For complex tables, you MUST allow multi-level nesting, even if there is only one hierarchy_key column.

5.1 Column-based hierarchies
- When there are multiple hierarchy_keys across different columns, you should:
  - For each row, create or move into a nested path defined by those columns in order (e.g. Region > City > Store).

5.2 Row-based block hierarchies (very important)
When the analysis_reason indicates that the hierarchy is primarily row-based or block-structured:
- Treat rows whose hierarchy_key cell is non-empty but whose value_leaves are all empty or missing as PARENT or GROUP rows (block headers).
- The data rows that follow a parent row are CHILD rows of that parent.
- In the JSON tree:
  - Create a top-level key for each parent row using the hierarchy key rule.
  - Under that key, create nested keys for each child row.

6. No Data Omission
- You must not omit any cell content from the input table.
- Every non-header row must contribute at least one object into the JSON tree.

7. Prohibited Behaviors
- Do NOT invent new textual labels that do not appear in the original table or headers.
- Do NOT rename, summarize, or translate headers or cell values.

[Input Table]:
{TABLE_AS_JSON_STRING}

[Normalized Headers]:
{NORMALIZED_HEADERS_FROM_STEP_1}

[Hierarchy Definition]:
{HIERARCHY_DEFINITION_FROM_STEP_2}

Your output must be a single, valid JSON object that fully represents the tree structure of the entire table.
Ensure that all data types (numbers, strings) are correctly preserved.
Do not output any other text or explanation.
\end{lstlisting}

\begin{lstlisting}[caption={Prompt Template for Foundation Models}, label={lst:foundation_prompt}]
You are an expert table analyst. Your task is to answer the question based on the provided table.

# Instructions:

1. Please enclose your final answer in square brackets, e.g., [Answer].
2. **Pay strict attention to the required answer type.**
   - If the question starts with **"Which"** or **"What"** (e.g., "Which item?", "What category?"), your answer must be a *name or description* (e.g., "Sell Product A"), **NOT** the associated numerical value.
   - If the question starts with **"How much"**, **"What is the value"**, or **"How many"**, the answer must be a *numerical value* (e.g., "120000").
3. **ONLY when** the answer is numerical (as per rule 2), you must check for any associated "units" (e.g., "ten thousand", "million") and provide the final converted numerical result.

## Table:
{table_str}

## Question:
{question}

## Final Answer:
\end{lstlisting}

\end{document}